\documentclass[review]{elsarticle}

\usepackage{lineno,hyperref}
\usepackage{graphicx}
\usepackage{caption}
\usepackage{float}
\usepackage{amsmath}
\usepackage{mathrsfs}
\usepackage{comment}
\usepackage{subcaption}
\modulolinenumbers[5]

\journal{Journal of \LaTeX\ Templates}

\begin{document}

\begin{frontmatter}

\title{A novel framework for generalization of deep hidden physics models}

\author{Vijay Kag}
\author{Birupaksha Pal\corref{mycorrespondingauthor}}
\address{Bosch Research and Technology Center, Bangalore}
\cortext[mycorrespondingauthor]{Corresponding author}
\ead{Birupaksha.Pal@in.bosch.com}

\address{123 Industrial layout, Koramangala, Bangalore 560095}

\begin{abstract}

Modelling of systems where the full system information is unknown is an oft encountered problem for various engineering and industrial applications, as it's either impossible to consider all the complex physics involved or simpler models are considered to keep within the limits of the available resources. Recent advances in greybox modelling like the deep hidden physics models address this space by combining data and physics. However, for most real-life applications, model generalizability is a key issue, as retraining a model for every small change in system inputs and parameters or modification in domain configuration can render the model economically unviable. In this work we present a novel enhancement to the idea of hidden physics models which can generalize for changes in system inputs, parameters and domains. We also show that this approach holds promise in system discovery as well and helps learn the hidden physics for the changed system inputs, parameters and domain configuration.
\end{abstract}
 

\end{frontmatter}

\section{Introduction}

In our modern software driven world, system modelling lies at the heart of most modern engineering and scientific innovation. System models are utilized in a wide range of applications, be it for predictive control and maintenance of two-wheeler engines or to develop robust digital twins for complex processes involving multiple connected machines. Generally, one would approach building a model using the available physical knowledge of the system. Use of the first principles greatly aid in providing deep insights into the working of the system and in model interpretability. But often it's either impossible to consider all the complex physics involved or simpler models are considered to keep within the limits of the available resources. This renders the considered model incomplete or deficient. Hence the best of classical numerical solvers will not be of any help as the accuracy of the model will be strictly limited by the scope of the available physical information. On the other hand, models built purely using data are extremely flexible and doesn't require any prior system knowledge. But they lack in utilization of the available system knowledge and suffers from low interpretability and generalizability. Nowadays with the buzz around explainable AI, the problem of interpretability is of great consequence and more so for safety critical systems.

This area of scientific machine learning (SciML) has a great deal of overlap with system identification, which in the recent past had multiple developments like the highly popular Sparse Identification of Non Linear Dynamics (SINDy methods) \cite{SINDY}. One challenge for these methods is in obtaining the derivatives which can be very sensitive to noise, though some new adaptations \cite{SINDYPI,EnsembleSINDy,DataDiscoverySINDY} promises faster convergence due to parallelism and also better handling of the noisy data. However, if the parameters of the equations are non-constant or the candidate terms are not linearly related then these sparse regression technique falls short. Next there are symbolic regression based methods which can circumvent some of these issues \cite{PhysRevE.94.012214,Symbreg1}, but one crucial challenge which remains with any symbolic regression based methods is that they are highly resource intensive. One point to be noted here is that all system identification methods involve identifying the underlying mathematical form of the system. To find the solution one must undertake the additional step of solving the discovered equations. Some other interesting approaches in the literature which aims to look for the solution all the while taking care of the missing parts of the governing equation includes \cite{HUANG2020109491}, here a grey box approach is shown where FEM has been used to discretize the physical system and a neural network has been used to replace the constitutive relation which is then trained on global response information. In \cite{ROBINSON2022333} a neural network based approach has been introduced where partially known system information is injected in intermediate layers to improve model accuracy. Another interesting new approach for system modelling where the full governing equations are not known called the hidden physics models. This is a new paradigm of methods where the idea was first introduced in \cite{RaissiDHP} as an architecture based on a pair of neural networks, it was named `deep hidden physics models' (DHPM) but core idea has be implemented by a pair of gaussian processes as well \cite{RAISSI2018125}, where the framework has been named `hidden physics models' (HPM) or by a combination of neural network and sparse regression\cite{Chen2021}, where it's called `PINN-SINDY' and further by a combination of symbolic regression and neural network as well. 
These methods which we shall go into details in the next sections involves enriching the cost function to include the information from the partially known equation and further makes use of automatic differentiation to evaluate the derivative of candidate terms for the hidden part. These are robust methods which can be used for both forward as well as inverse modelling \cite{8970456, ZHANG2021}. Infusing system knowledge in cost function for neural network training exists in the world of physics informed neural networks (PINNs) \cite{raissi2017physics,RAISSI2019686, PINNreview} as well, but the main difference with hidden physics models lie in the usage of a second network to account for the unknown physics. The single network PINN can be used as an effective surrogate modelling technique if the full governing equations are known, but the single network can't account for the unknown physics and would be limited by the available system information. 

Apart from the problem of hidden physics, it might not always be possible to train the model for all possible inputs, parameters and domain configurations. Hence, model generalizability poses an important practical challenge which has direct consequence in model deployment. This makes the problem of building system model for real life engineering and industrial applications trickier.

Here in this paper, we shall propose a novel modification to the idea of hidden physics models and develop a model architecture by which we wish to target three aspects. First, the model should be able to handle situations where the governing equations are not fully known. Second the model should be able to react to changes in system like inputs, parameters and domain configuration and combinations of them. Third, we wish to make sure the interpretability aspect of the built models is also well taken care of. Hence, we expect the trained model should be able to learn adequately the hidden physics of the system as well. 

The proposed methodology is based on a pair of neural networks (DHPM). Neural networks are extremely versatile and flexible in learning a varied kind of systems and are extremely simple and easy to implement. Moreover, the proposed framework can be extended for gaussian processes, sparse and symbolic regression as well. 
 Though there are multiple methods to handle systems with partially known physics, to the best of the authors' knowledge the generalizability aspect we mention here has not been explored earlier, more so in conjugation of all the three aspects mentioned earlier.

The paper has been structured into 6 sections. Following the introduction in section 2 the core idea of deep hidden physics models has been described. In section 3, the proposed architecture for generalization of inputs has been described along with detailed results and plots. In Section 4, the methodology for input generalization has been extended to include both parameter and input change, along with the results. In section 5, a methodology for generalization of  input and domain together has been introduced with it's corresponding results and finally in section 6 we conclude by summarizing the paper.

\section{Deep Hidden Physics Models}

In this section we introduce the basic structure and idea of deep hidden physics models.

We consider a general form of  partial differential equation represented by 
\begin{equation}\label{deficient_eq}
	\frac{\partial u}{\partial t} + \mathcal{N}(x,t,u,u_x,u_{xx},...)=0
\end{equation}

Here, $u$ represents the state or solution and  $\mathcal{N}$ represents the hidden dynamics. To keep things generic we consider only the time derivative as known. This is also a fair assumption as for most  situations one would atleast know whether the system is a steady or a transient one. Now, for $N$ available measurement data of the system $u_d(x_i,t_i)$, for $i=1 \  \text{to} \ N$,  one can build a regression model for $u$ based on coordinates $x,t$ as input features with the following cost function.

\begin{equation}\label{Loss_data}
\text{Loss data =}   \frac{1}{N} \sum_{i=1}^{N} |u(x_i,t_i)-u_d(x_i,t_i)|^2 
\end{equation}
 
This cost function as derived using measurement data will be referred to as data loss. Next, the work that remains would be to optimize/train the parameters of the chosen regression model to minimize the data loss. As can be clearly seen, the exercise till here can give a trained model though the solution completely disregards any available physical information of the system. The idea of DHPMs \cite{RaissiDHP} is a neural network based architecture which incorporates the available physical information into the cost function, over and above the data loss.

\begin{figure}[H]
 	\centering
 	\includegraphics[width=5cm]{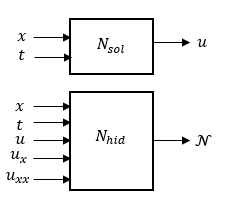}	
 	\caption{ Networks for the solution and the hidden physics}
 	\label{fig:HPM_net_archt}
 \end{figure}

\begin{figure}[H]
	\centering
	\includegraphics[width=\textwidth,height=4.5cm]{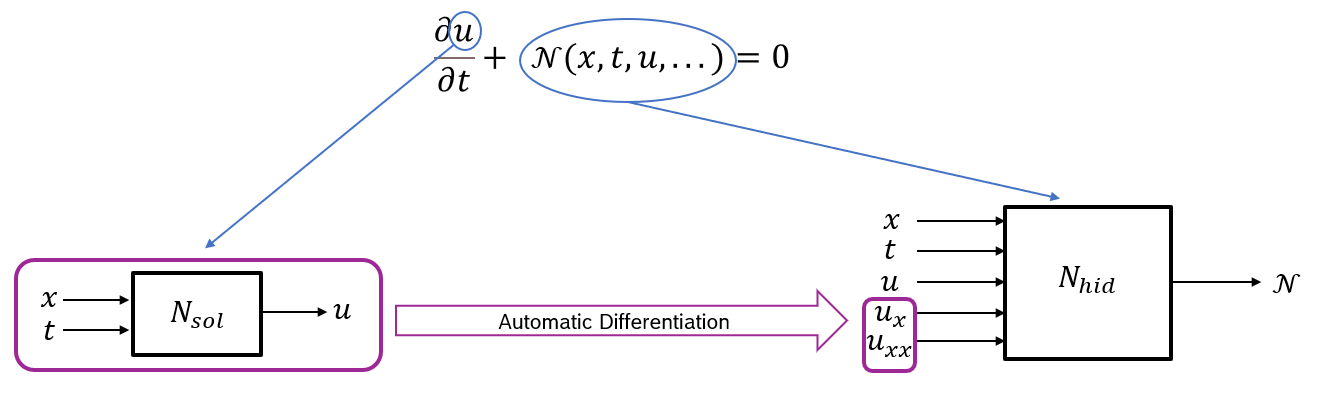}
	\caption{ Deep hidden Physics model framework}
	\label{model_framework_explained}	
\end{figure}

 The strategy consists of two neural networks as shown in Figure  \ref{fig:HPM_net_archt}. 
Here, the network $N_{sol}$ is an approximation of the state $u$ and the network $N_{hid}$ as an approximation of the hidden part of the governing equation denoted by $\mathcal{N}$.
It must be noted here that the number of candidate terms used for the network $N_{hid}$ is user defined and based on domain knowledge. It makes sense, because even if the constitutive relations are unknown still a domain expert can have a fair idea of the approximate constitutive terms which affect the dynamics of the system under study. 
In the schematic \ref{model_framework_explained} for explanation we have shown spatial derivatives of the state up to 2nd order i.e. $(u_t,u_x,u_{xx},..) $. Usage of higher order derivatives as candidate terms is a fairly well-known technique for system identification. Now, as in most situations it may not be possible to measure these derivatives hence numerical approximations are commonly used, but there the problem is that numerical derivatives can show high oscillations for noisy data and lead to inaccurate results. Here, this problem is easily circumvented by utilizing the autograd feature of neural networks.

Moreover, as we have two neural networks now, we need a way to connect them together. This can be brought about by a neat trick of utilizing the form of the system equation. In doing so we will observe that not only will this help connect the two networks but additionally it would embed the information from physics into the solution. 

We define the residual of PDE say $g$ as the following, 
\begin{equation}
g(x,t) = \frac{\partial u}{\partial t} + \mathcal{N}(x,t,u,u_x,u_{xx},...)
\end{equation}

It must be noted that for the equation \eqref{deficient_eq} to be satisfied, this residual where $u$ and $\mathcal{N}$ are approximated by the networks $N_{sol}$ and $N_{hid}$ should be approximately 0 at any point in the domain. 

The parameters of networks $N_{sol}$ and $N_{hid}$ can be learned by minimizing Total Loss as defined,
\begin{equation}\label{def_Total_Loss}
\text{Total Loss = Loss data + Loss equation } 
\end{equation}
where Loss data is defined in \ref{Loss_data} , and `Loss equation' defined as follows
\begin{equation}\label{Loss_eqn}
\text{Loss equation =}  \frac{1}{M} \sum_{i=1}^{M} |g(x_i,t_i)|^2 
\end{equation}

Here, the $M$ points considered are known as collocation points, and they are used to constrain the model to follow the governing equations. These points need not be measurement data but can be any random points in the domain. 
 Once both the networks are trained, the network $N_{sol}$ represents the solution state and the network $N_{hid}$ represents the hidden part of the governing equation or $\mathcal{N}$  in an implicit black box form. 
 
 It must be noted here that, in purely system identification schemes an additional step for solving the obtained equations have to be performed. Whereas, this method provides both the solution state as well as the representation of the hidden physics, and thus can be seen to combine the benefits of system identification along with simulation schemes.
 Further, the cost function utilizes both the partially available physical knowledge the data. Hence, the solution obtained is more explainable as compared to only using measurement data and this is a crucial differentiator from traditional regression models like multilayer perceptron (MLP) for that matter. With, the recent focus in the research community as well as industry towards more explainable AI, this can act as an important USP. Moreover, as we use information from physics hence the amount of data required for modelling can be lower than traditional data-based methods.

\textbf{Generalization for system change}

A key challenge in terms of modelling faced in industrial applications is the capacity for generalization of models to react to system change. Examples of system change could be something like a new type of input or loading function that needs to be tested, or change in some system parameter due to change in materials or an alteration or deformation in the domain. As most systems and processes are well integrated into tight pipelines, hence re-measuring and collection of data corresponding to small changes is a difficult and can even render the modelling and virtualization task economically unviable. However, if newer data corresponding to the altered system specification isn't considered and the model not retrained that might lead to less than acceptable level of accuracy. Hence, a desired property of a system model for industrial applications is it's ability to generalize for changes in inputs, parameters and domain and combinations of them. 
In the following sections we delve into this aspect and introduce novel improvements in the architecture of DHPMs to account for the system changes. 

\section{Generalization for inputs}\label{inp_gen_sec}
We first consider generalization in terms of change in input functions. In this regard it's important to note that in recent years, development in the area of operator learning aims to target this issue. There are some popular operator learning methods like Deep operator networks (DeepONet) \cite{DeepONet1,GOSWAMI20221}, Laplace neural operator (LNO) \cite{cao2023lno,chen2023learning} and Fourier neural operators (FNO) \cite{li2021fourier, FNO2}. All of them are purely data-based approaches which doesn't explicitly use any physical information of the system. Here in this work, we present a novel architecture based on the previously introduced DHPM for input generalization, which utilizes both the measurement data and the partially available system information.  

\begin{figure}[H]
	\centering
	\includegraphics[width=6cm]{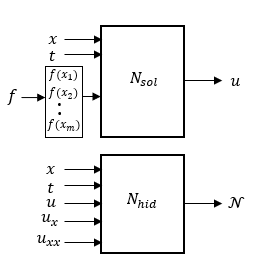}
	\caption{Network architecture for generalization of input}	
	\label{HPM_gen_input_net_archt}
\end{figure}

The strategy we adapt is to enrich the input feature space for $N_{sol}$ by adding discretized form of the input functions. This is done to incorporate the effect of the inputs within the network during training. Our expectation is that once trained for a sufficient number of input functions the model can learn to generalize for unseen inputs. As shown in figure \ref{HPM_gen_input_net_archt}, we modify the network $N_{sol}$ to add the input function $f$ discretized at $m$ sensor points in addition to the existing spatio-temporal variables.

Next, the loss function is created as shown in \ref{def_Total_Loss}. It should be noted that both the equation and data losses must correspondingly be modified so as to account for the multiple input functions used for training.

We showcase the results of the proposed strategy for a reaction-diffusion system. The reaction-diffusion (RD) equation models several physical phenomenoa in nature \cite{Kop04}. It describes the spatio-temporal evolution of chemical concentration for one or more substances. 

The equation is given by the following, 
\begin{equation}\label{reaction_diffusion_eqn}
\frac{\partial u}{\partial t} = D \frac{\partial^2 u}{\partial x^2} + K u^2
\end{equation}
Here $D, K$ represent diffusion coefficient and reaction rate respectively. We consider the domain $x \in [0,L]$ and $t \in [0,10]$, with initial condition as  $u(x,0)=f(x)$ and boundary condition as $u(0,t)=u(L,t)=0$.

To check the proposed methodology of generalized DHPM we shall consider only the time derivative term in the RD equation to be known and the rest hidden. We will examine the capacity of the proposed framework to accurately predict the solution state of the RD equation by utilizing the information of the time derivative and some data. The data which we use as measurement data, is  obtained by solving the RD equations for different input functions in a $(x,t)$ domian.

To simulate the RD system we shall consider different initial conditions given as \ref{periodic_input_function}. The nature of functions are chosen such that it automatically satisfies the boundary condition of the problem.
 
\begin{equation}\label{periodic_input_function}
f(x) = \sum_{k=1}^{N_f} A_k sin(k \pi x /L)
\end{equation}  
The coefficients $A_k$'s  are uniformly and randomly generated, we further consider, 
$$
|A_k |\leq 0.4,   N_f=5, L=1
$$

 We solve the RD equations using finite difference with forward time and centered space scheme (FTCS) with $\delta x = 5 \times 10^{-3}$ and   $\delta t = 10^{-3}$. We obtain the solution at 201 $\times$101 equispaced spatio-temporal grid points corresponding to each input function.   
 
\begin{figure}[H]
	\centering
	\includegraphics[width=10cm]{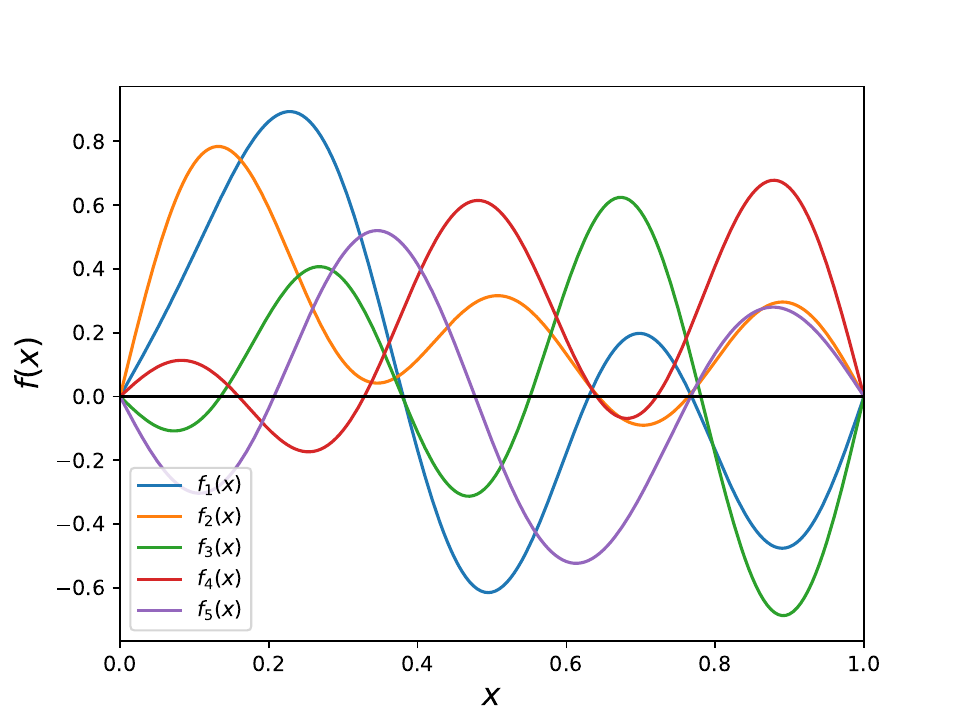}
	\caption{A few random input functions }	
	\label{input_function_plot}
\end{figure}

For training the network architecture \ref{HPM_gen_input_net_archt}, we generate $n_{fun}$ number of different input function as given in the form \ref{periodic_input_function} by changing the coefficients $A_k$'s. The figure \ref{input_function_plot} shows the plot of a few random input functions. Note that value of input function should be zero at the boundary i.e. $f(0)=f(L)=0$ so as to satisfy the given boundary condition as mentioned above.  

We pass input function as input features discretizing each at $m=50$ equi-spaced points called the `sensor points'. We build a vector of the discretized points and concatenate it with the variables $x$ and $t$. Hence, the input to the network $N_{sol}$ is now a vector with 52 components, 50 for the discretized input function $f$, and 2 for the space and time variables. The architecture we consider for $N_{sol}$ is [52,100,100,100,1], i.e. 3 hidden layers with 100 neurons each. Next, we have assumed that the quantities $(x,t,u,u_x,u_{xx})$ influence the hidden dynamics which is represented by the network $N_{hid}$. Hence, with these 5 input features, the architecture of $N_{hid}$ has been considered as [5,100,100,100,1]. Further, we have considered the parameters $D, K $ to be $10^{-3}$ and $10^{-3}$  respectively. 

 We consider the number of input function for training to be $n_{fun}$ and the number of $(x,t)$ grid points where the state is evaluated/measured to be $n_{data}$, thus the total number of training data points are $n_{train} = n_{fun} \times n_{data}$. The training data is splitted into multiple minibatches. We have considered the batch size i.e., the number of data points in each minibatch to be $n_{data}$ so that within the batch, the data points corresponds to same input function and therefore the total number of batches are $n_{train}/n_{data} = n_{fun}$. Training across entire batches constitute a single epoch and the model is trained for multiple epochs. The number of collocation points at each gradient descent steps considered to be 5000. These collocation points are randomly sampled using Latin-Hypercube sampling (LHS) method \cite{LATINHYCUBE}. To minimize the Total Loss, we use gradient based optimizer, ADAM. The optimization algorithm is described as follows. 	
\begin{align*}
For ( & epoch=1, epoch \leq n_{epoch},epoch++) \{ \\
&For (i=1,i \leq n_{fun},i++)\{ \\
& \hspace{1cm} \text{1. Randomly select collocation points and calculate Loss equation}  \\
&\hspace{1cm} \text{2. Calculate Loss data, MSE at  $(x_j,t_j)$ for $ 1 \le j \le n_{data}$  }  \\
&\hspace{1cm} \text{3. Compute Total Loss =  Loss  data + Loss equation} \\
&\hspace{1cm} \text{4. Update networks parameters by minimizing Total loss} \\
\}\}
\end{align*}

The model is trained with 1000 epochs with learning rate of $10^{-3}$ and the next 1000 epochs with learning rate of $10^{-4}$.  
Further, to find the optimal number of input function $n_{fun}$ for training and number of measurement points $n_{data}$, the network has been trained with two sets of $n_{fun} = \{100,200\} $ and 5 sets of $n_{data} =\{100,500,1000,2000,4000\}$ that accounts for  approximately $  \{0.5 \%,2.5 \%,4.9\%,9.9\%,19.7\% \}$ of entire simulation data. For the error measurement we use the metric, relative $L_2 \  \text{error}$ or simply denoted as $L_2$ Error given by \ref{l2_error_def},

\begin{equation}\label{l2_error_def}
L_2 \ \text{Error} =  \frac{\lVert u^* - u\rVert_2} { \lVert u^* \rVert_2 }
\end{equation}
,where $\lVert \cdot \rVert_2$ denotes $L_2$ norm, also  $u^*, u$ denotes actual and predicted values.

\begin{figure}[H]
	\centering
	\includegraphics[width=10cm]{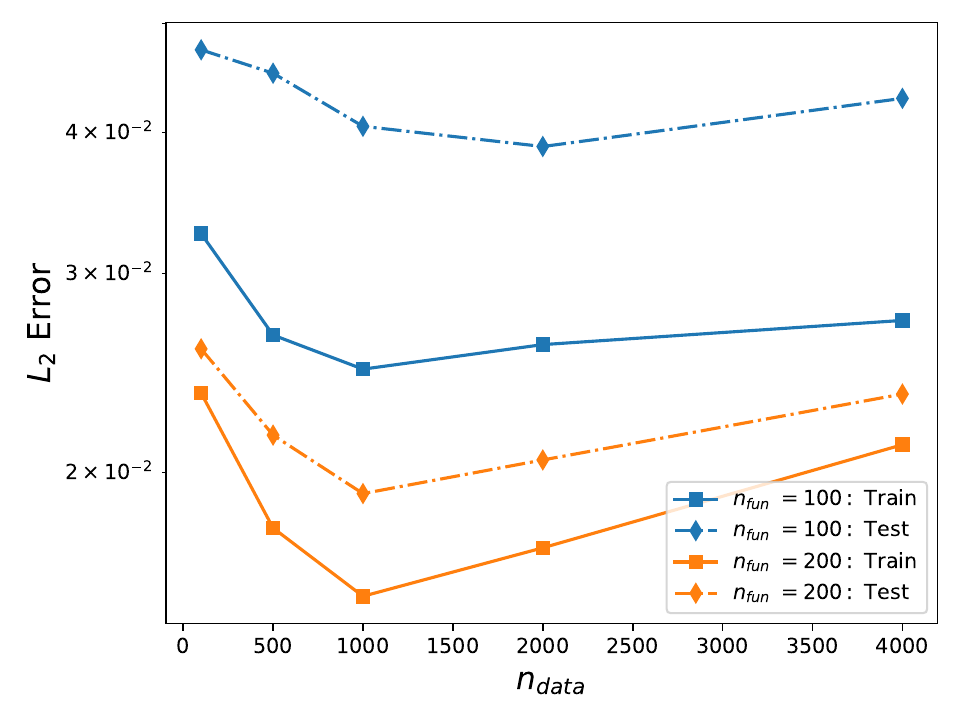}
	\caption{Error corresponding to number of data points, The train error is calculated by taking mean of $L_2 \  \text{Error}$ of solutions predicted for $n_{fun}$ number of input function that were used for training and similarly the test error is calculated by taking mean of  $L_2 \  \text{Error}$ of solutions predicted for 1000 random input functions of the same form given in \ref{periodic_input_function}.}	
	\label{train_test_data}
\end{figure}

\begin{figure}[H]
	\centering
	\includegraphics[width=10cm]{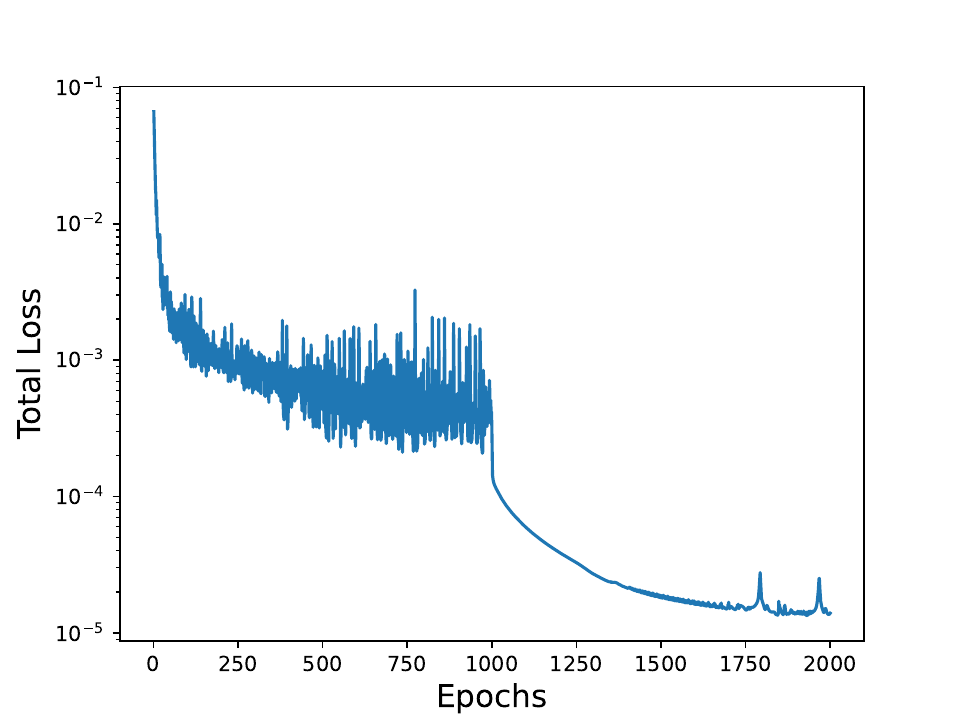}
	\caption{Trajectory of Total Loss over the training process for optimal model}	
	\label{total_loss_plot}
\end{figure}

\begin{figure}[H]
\centering
\begin{subfigure}{0.4\textwidth}
    \includegraphics[width=6cm]{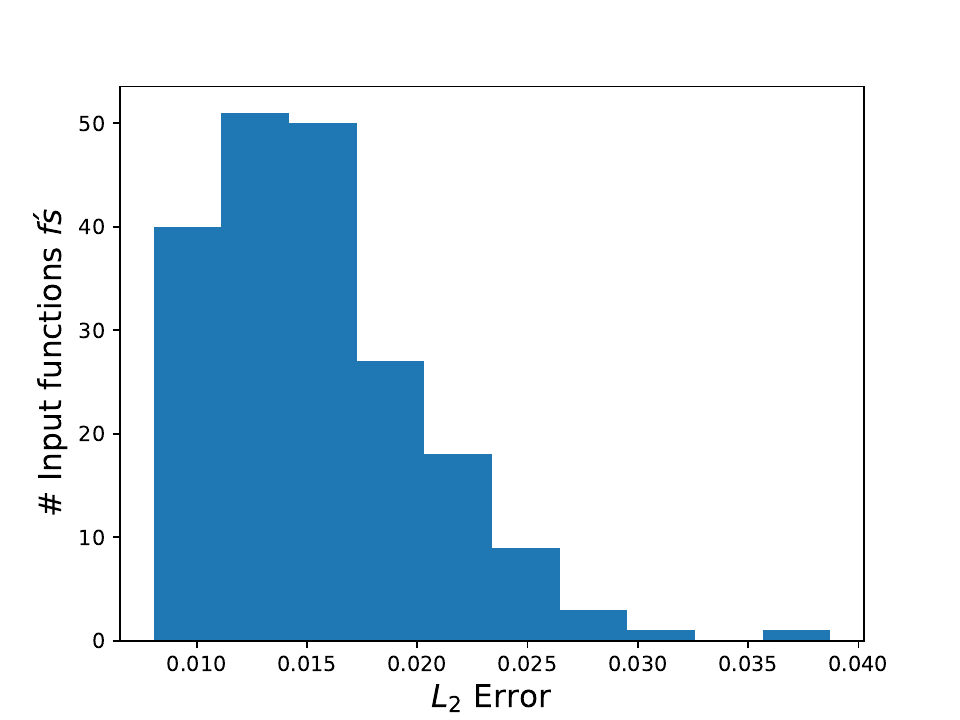}
    \caption{Distribution of error for predicted solutions from 200 trained input functions, mean  $1.55e-2$ and std. deviation  $4.93e-3$ }
    \label{train_error_distribution}
\end{subfigure}
\hfill
\begin{subfigure}{0.4\textwidth}
    \includegraphics[width=6cm]{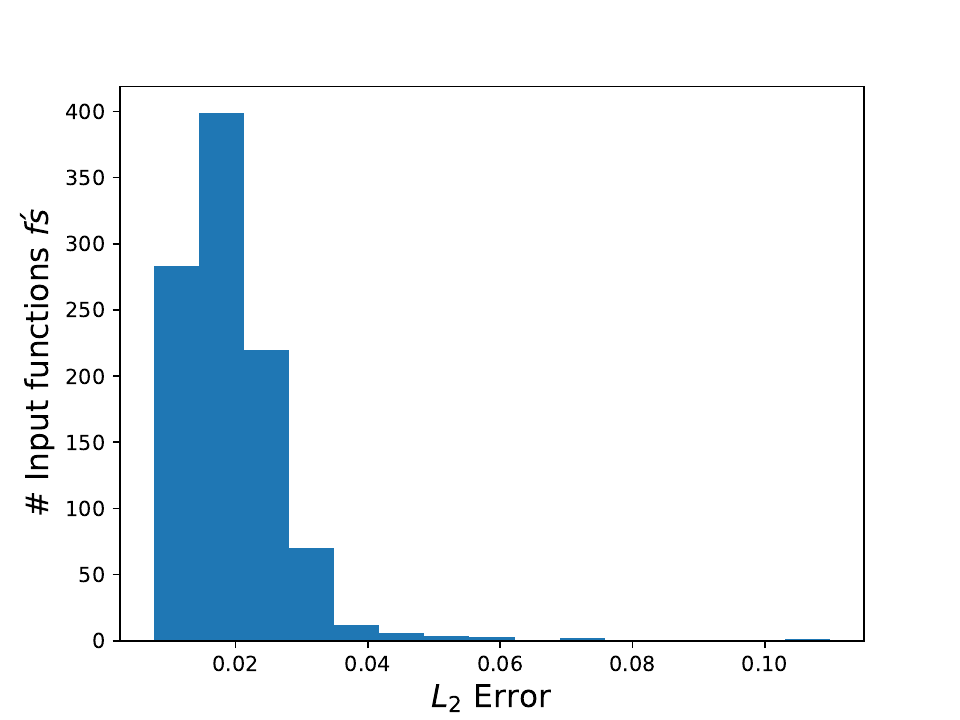}
    \caption{Distribution of error for predicted solutions from 1000  random input functions,  mean $1.91e-2$ and std. deviation  $8.04e-3$ }	
    \label{test_error_distribution}
\end{subfigure}
\caption{Distribution of train and test error}
\label{train_test_dist_error}
\end{figure}

Once the model is trained, we predict the solution at all the spatio-temporal grid points, numbering $201 \times 101$ and then calculate $L_{2}$ error. As we can observe in Figure \ref{train_test_data}, the minimum $L_{2}$ error of $1.91e-2$ is achieved for $n_{fun} = 200 $  and $n_{data} = 1000$ i.e., $4.9\%$ of simulation data. By increasing more training data, we don't find any significant improvement in the error and conclude that a bigger network might be required to utilize higher amount of data. Therefore, we find this to be an optimal model with relatively small error with a small amount of training data. The decay of loss during training for the optimal model is shown in figure \ref{total_loss_plot}. The plot in figure \ref{train_test_dist_error} shows the distrubution of train and test error for the optimal model. As we can observe that the mean $L_2 \  \text{Error}$ for the train, test are $1.55e-2$ and $1.91e-2$ respectively with standard deviation of $\mathcal{O}(10^{-3})$. As, there isn't much of a difference in train and test errors (both are of similar order of magnitude), therefore, we we can say that the model is not overfitting to the training data. 

\subsection{Prediction for similar kind of unseen input function}

\begin{figure}[H]
	\centering
	\includegraphics[width=\textwidth,height=3cm]{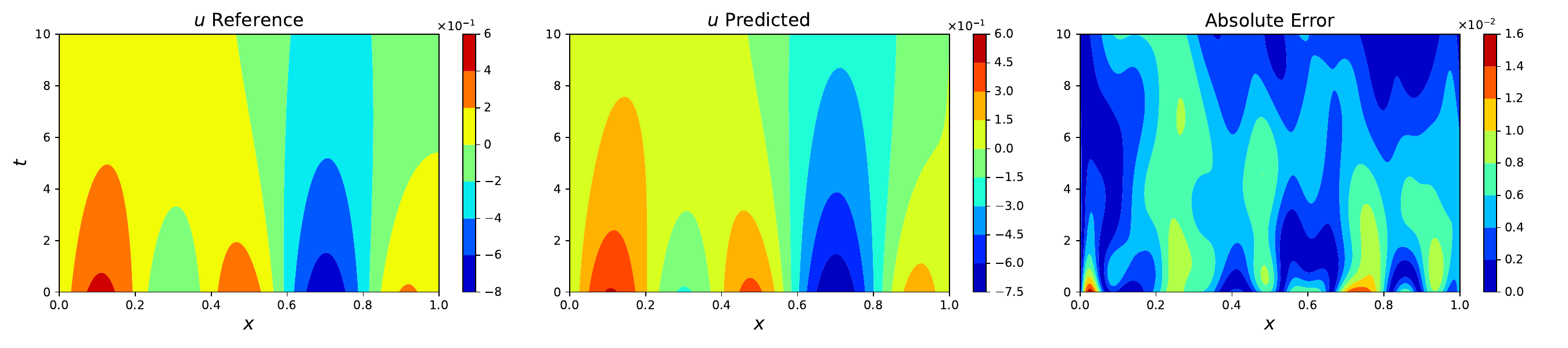}
	\caption{Reference solution vs model prediction, $L_2 \  \text{Error} \ 2.37e-2$ }	
	\label{input_functions_samekind_state_prediction}
\end{figure}

We predict the solution for a random periodic input function of the form as given in \ref{periodic_input_function}, but with unseen coefficients. The comparison of the reference and the predicted state is shown in figure \ref{input_functions_samekind_state_prediction}.

\begin{figure}[H]
	\centering
	\includegraphics[width=\textwidth,height=3cm]{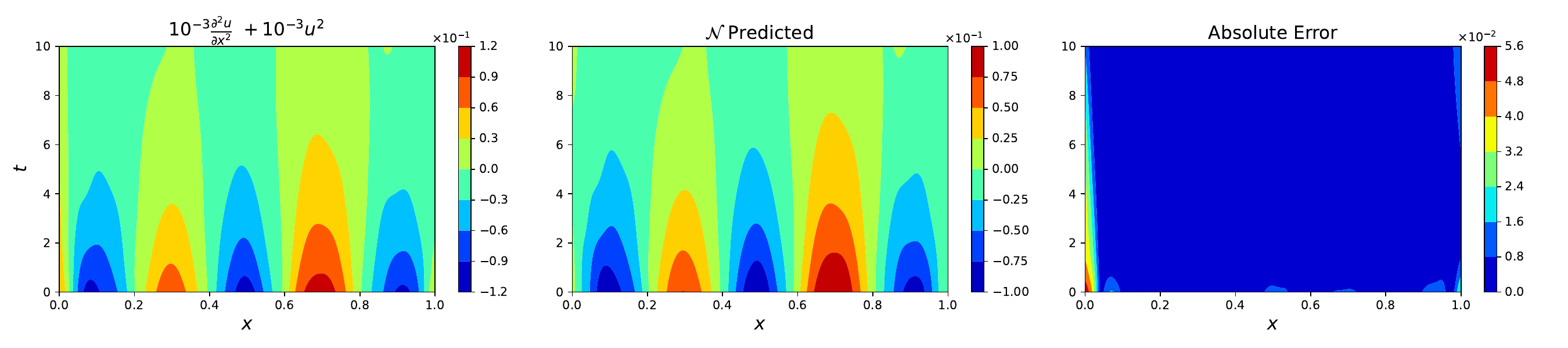}
	\caption{Comparison of actual hidden PDE terms and the network $N_{hid}$'s output $\mathcal{N}$ }	
	\label{input_functions_samekind_hidden_prediction}
\end{figure}

The plot shows a close approximation of the prediction with the reference data. Further, we also study the output $\mathcal{N}$ from network $N_{hid}$ that is an approximation of hidden part of equation and gives a blackbox/implcit representation of the hidden physics.  We first predict $u$ from network $N_{sol}$ and then calculate the actual terms in PDE i.e., $ 10^{-3} \frac{\partial^2 u}{\partial x^2}+ 10^{-3} u^2$. Here, the gradient are obtained using automatic differentiation technique. Note that in the subsequent sections this same method for evaluating and comparing of the hidden terms have been used. The contour in figure \ref{input_functions_samekind_hidden_prediction} shows the comparison of hidden physics predicted by the model vs the reference. 

.

\subsection{Prediction for out of distribution input functions}
To further check the capacity of input generalization we predict the solution for different kind of input functions other that the periodic ones as shown in \ref{periodic_input_function}. 

\subsubsection{Quadratic function}
\begin{figure}[H]
	\centering
	\includegraphics[width=\textwidth,height=3cm]{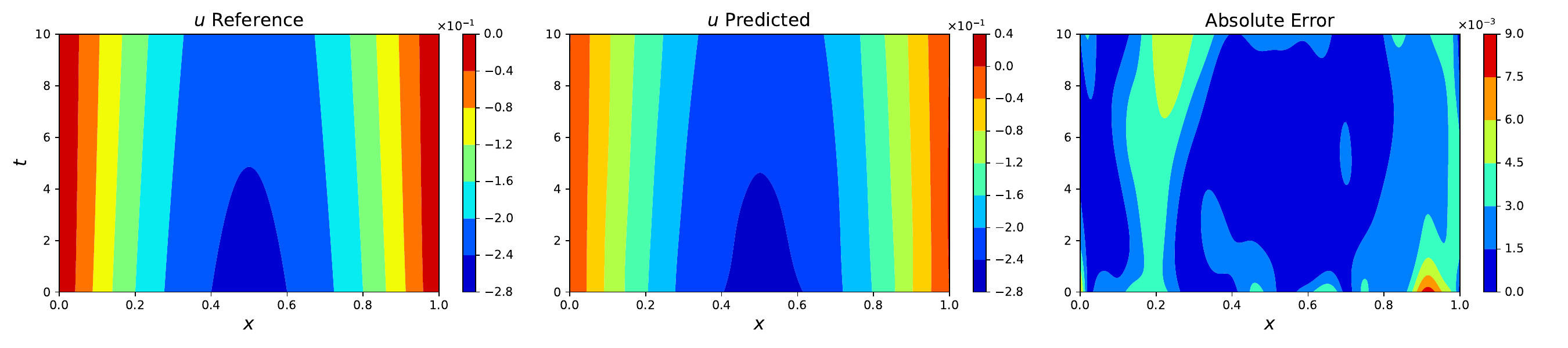}
	\caption{Reference vs model prediction for input $f=x^2-x$ ,  $L_2 \  \text{Error} \ 1.25e-2$ }	
	\label{quadratic_input_gen_state_prediction}
\end{figure}

First we consider a quadratic function of the form $f(x)=x^2-x$. The comparison between the trained model prediction and the reference solution is shown in Figure \ref{quadratic_input_gen_state_prediction}.

\begin{figure}[H]
	\centering
	\includegraphics[width=\textwidth,height=3cm]{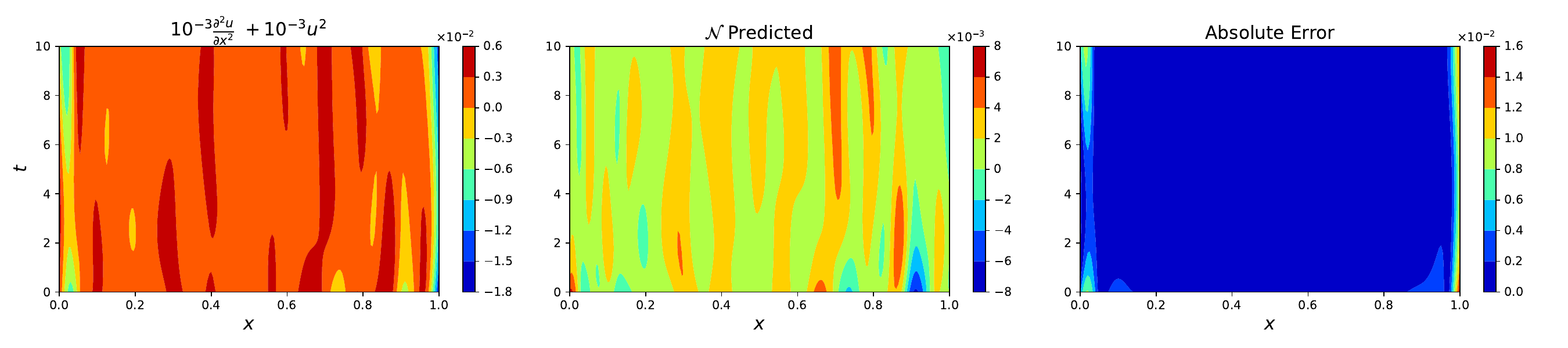}
	\caption{Comparison of actual hidden PDE terms and the network $N_{hid}$'s output $\mathcal{N}$ }	
	\label{quadratic_input_gen_hidden_state_prediction}
\end{figure}

The next plot in figure \ref{quadratic_input_gen_hidden_state_prediction} shows the comparison between the actual hidden terms of the PDE and the learned physics from $\mathcal{N}$. 

\subsubsection{Cubic function}
\begin{figure}[H]
	\centering
	\includegraphics[width=\textwidth,height=3cm]{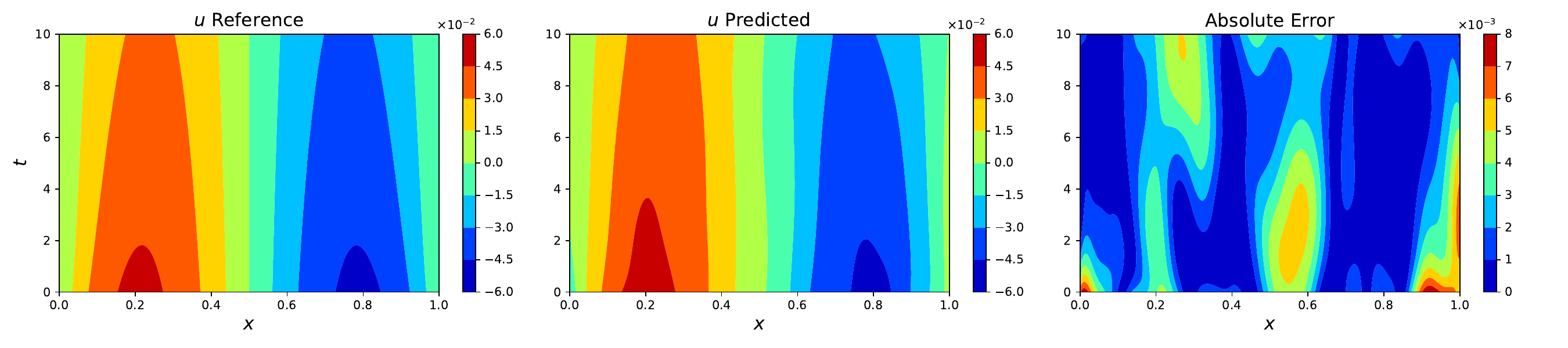}
	\caption{Reference vs model prediction for input $f=x^3-1.5x^2+0.5x$ ,  $L_2 \  \text{Error} \ 8.25e-2$ }	
	\label{cubic_input_gen_state_prediction}
\end{figure}

Next, we consider a cubic input function $f=x^3-1.5x^2+0.5x$.  Comparison between the model prediction and reference solution is shown in Figure \ref{cubic_input_gen_state_prediction}.

\begin{figure}[H]
	\centering
	\includegraphics[width=\textwidth,height=3cm]{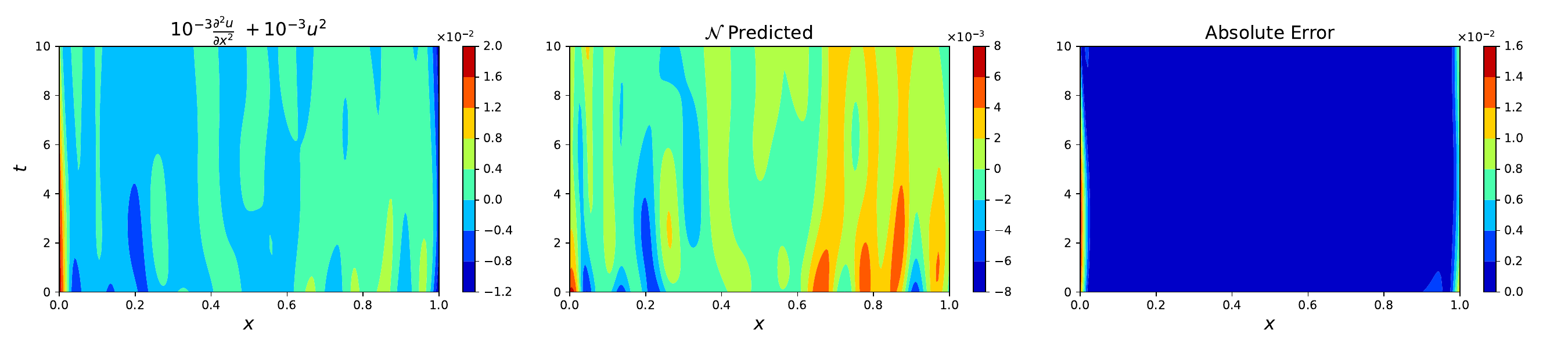}
	\caption{Comparison of actual hidden PDE terms and the network $N_{hid}$'s output $\mathcal{N}$ }	
	\label{cubic_input_gen_hidden_state_prediction}
\end{figure}

The next plot in figure \ref{cubic_input_gen_hidden_state_prediction} shows the comparison between the actual hidden terms of the PDE and the learned physics from $\mathcal{N}$.

\subsubsection{Trigonometric function}
\begin{figure}[H]
	\centering
	\includegraphics[width=\textwidth,height=3cm]{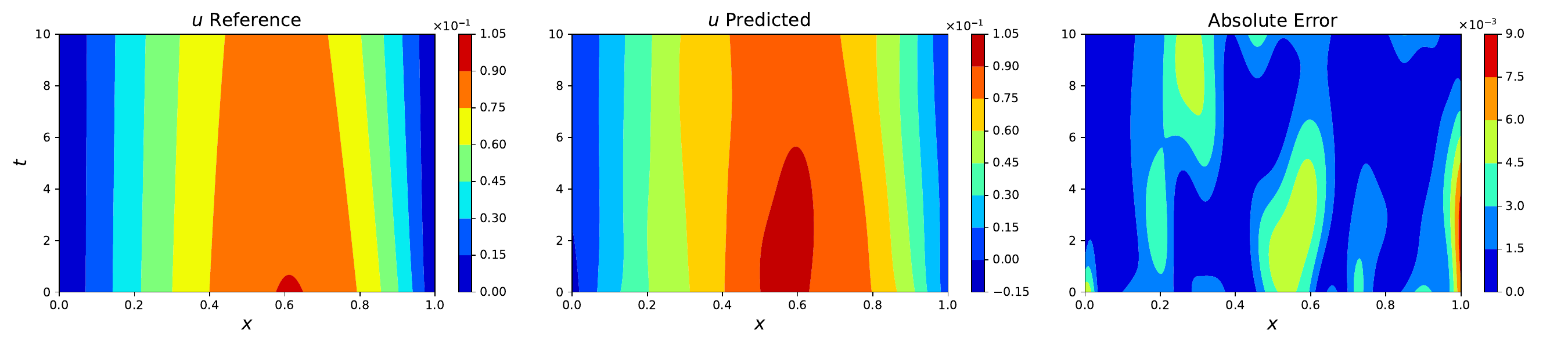}
	\caption{Reference vs model prediction for input $f=x-tan(\frac{\pi x}{4})$ ,   $L_2 \  \text{Error} \ 3.78e-2$ }	
	\label{trigono_input_gen_state_prediction}
\end{figure}

Finally, we consider a trigonometric function $f=x-tan(\frac{\pi x}{4})$. Comparison between the model prediction and reference solution is shown in figure \ref{trigono_input_gen_state_prediction}.

\begin{figure}[H]
	\centering
	\includegraphics[width=\textwidth,height=3cm]{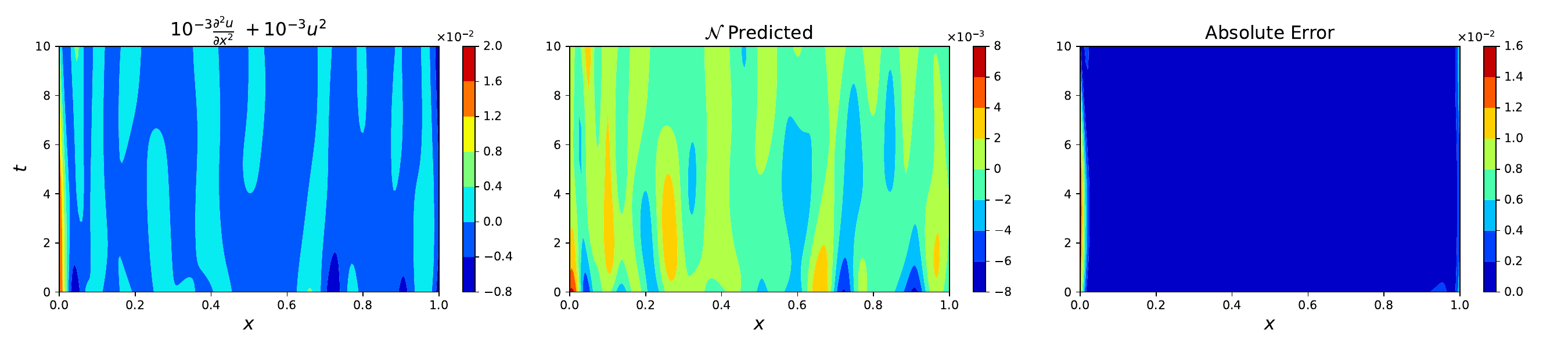}
	\caption{Comparison of actual PDE terms and output $\mathcal{N}$ }	
	\label{trigono_input_gen_hidden_state_prediction}
\end{figure}

Next, figure \ref{trigono_input_gen_hidden_state_prediction} shows the comparison between the output $\mathcal{N}$ with actual hidden PDE terms for trigonometric input.

These results showcase that even with one known term, that model is able to learn the effect of input change like an operator learning algorithm. Moreover, it holds promise in learning the hidden part of the equation too as in system discovery schemes albeit in a blackbox form.    

\section{Generalization of parameters}
In this section we increase the complexity of our problem by considering a situation where over and above input change the model would have to deal with change in parameters as well. A common example could be in the form of changes in the material properties of the underlying system. In this direction it's important to note here that in most real life examples for each parameter of interest there's a range within which it lies. Taking this cue, we extend the network architecture as described in figure \ref{HPM_gen_input_net_archt}, by enriching the input feature space of the network $N_{sol}$ by adding the system parameters. Our expectation is that if the model can be trained trained for a bunch of parameter values within a range, then the trained model should be able to learn the effect of the parameters within that range. We shall further elaborate on the architecture and the scheme in details in this section.
 
\begin{figure}[H]
	\centering
	\includegraphics[width=6cm]{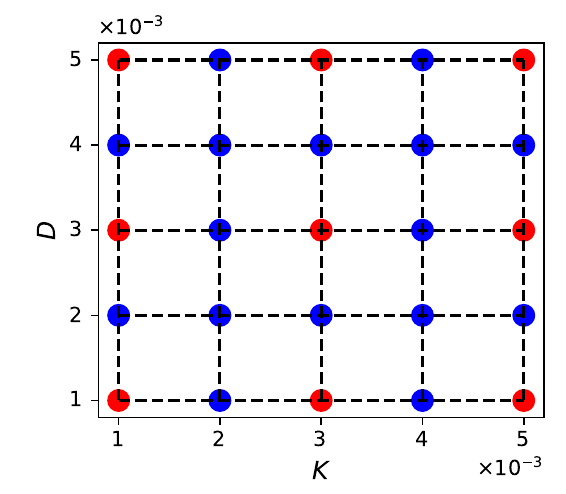}
	\caption{Diffusion coefficient ($D$) and reaction rate ($K$) considered, parameters in  `red', `blue' dot are training and testing respectively}	
	\label{D_and_K_plot}
\end{figure}

To demonstrate the proposed methodology, and to keep things consistent, we consider the same reaction diffusion system as given by  equation \ref{reaction_diffusion_eqn}. It can be noted the RD system has two parameters, the diffusion coefficient and reaction rate that is represented as $D$ and $K$ respectively. We train the model for a 3 $\times$ 3 combination of $D$ and $K$ with each having the values in set $\{10^{-3},3 \times 10^{-3},5 \times 10^{-3} \}$. The schematic of the considered combinations of $D$ and $K$ is shown in figure \ref{D_and_K_plot}.
 
\begin{figure}[H]
	\centering
	\includegraphics[width=6cm]{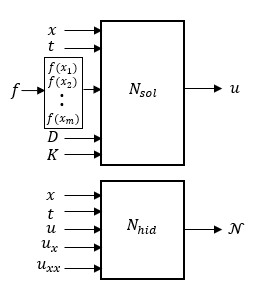}
	\caption{Network architecture for parameter and input function generalization}	
	\label{net_arch_parameter}
\end{figure}

Here, for the RD system only the time derivative term is considered as known. Further, both the parameters $D, K$ and the input function $f$ changes. We consider the two parameters to be additional inputs to the network $N_{sol}$ in addition to the discretized input function $f$ at 50 equispaced sensor points and the two spatio temporal variables $x$ and $t$. Further, for the network $N_{hid}$, we consider $(x,t,u,u_x,u_{xx})$ as the candidate terms. With these descriptions, the neural network architecture for network $N_{sol}$ is [50+4,100,100,100,1], and for the network $N_{hid}$ is [5,100,100,100,1]. Schematically it can be represented as shown in figure \ref{net_arch_parameter}.

\begin{figure}[H]
	\centering
	\includegraphics[width=10cm]{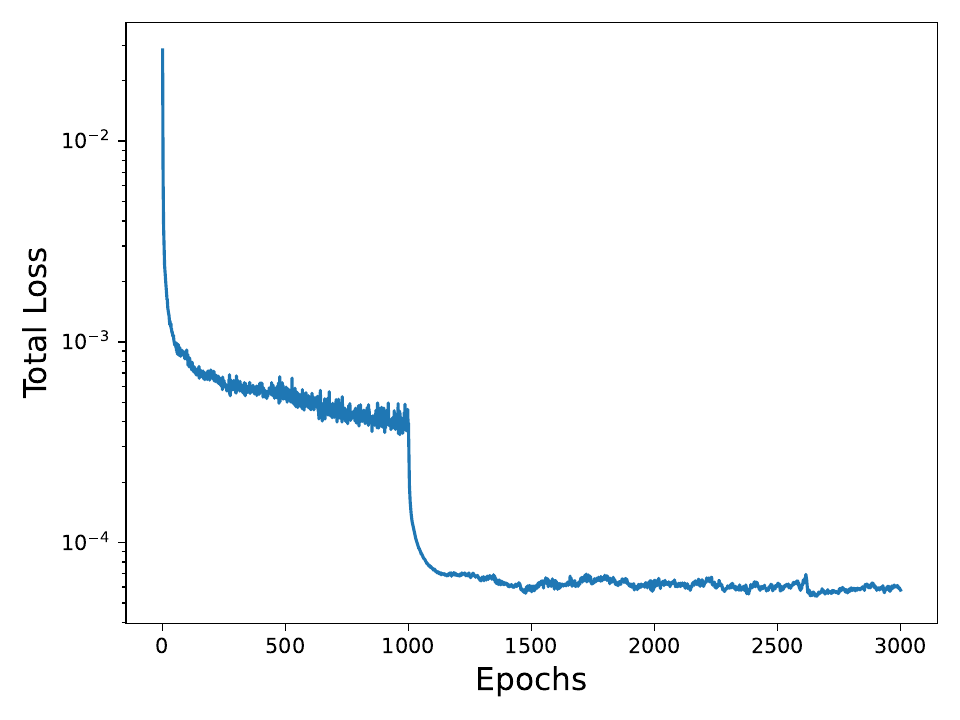}
	\caption{ Loss decay over the training epochs }	
	\label{total_loss_parameter_plot}
\end{figure}

 Moreover, we train the model with random input functions of the form given in \ref{periodic_input_function}. Further, we consider 200 random input functions for each of the 9 combinations of the $D$ and $K$ parameters for training. 
 We consider 1000 random collocation points at each gradient descent step and $n_{data} = 500$. The training data used is approximately $2.5\%$ of entire simulation data. The cost function is same as in equation \ref{def_Total_Loss}, with corresponding updates to account for the input fucntions and the parameters used in training. We train the model using Adam optimizer with learning rate $10^{-3}$ for first 1000 epochs and $10^{-4}$ for the next 2000 epochs. The loss decay over the training epochs is shown in figure \ref{total_loss_parameter_plot}, which shows convergence in the training.
 
 In the following subsections, we show the accuracy of the trained model in predicting for unseen inputs as well as unseen combination of $D$ and $K$. This will demonstrate the efficacy of the proposed methodology in learning the combined effect of parameter and input change. All the data of the RD system used for training and reference has been derived as shown in section \ref{inp_gen_sec}. 
 
\subsection{Unseen parameter and similar kind of unseen input function}
\begin{figure}[H]
 	\centering
 	\includegraphics[width=\textwidth,height=3cm]{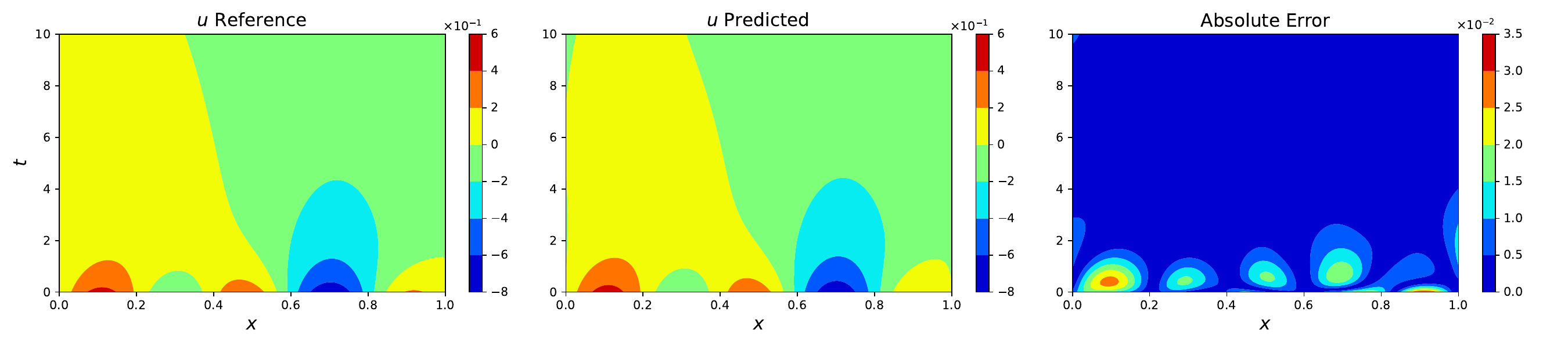}
 	\caption{Reference vs predicted $u$  for parameter  $D = 4 \times 10^{-3}$,  $K = 4 \times 10^{-3}$ and input $f$  as given in form  \ref{periodic_input_function} , $L_2 \  \text{Error} \ 3.35e-2$ }	
 	\label{prediction_same_kind_parameter_gen_soln}
 \end{figure}

We first predict for an unseen random periodic input function $f$ as given in form \ref{periodic_input_function}  and unseen parameters value of $D = 4 \times 10^{-3}$ and $K = 4 \times 10^{-3}$. The plot in Figure \ref{prediction_same_kind_parameter_gen_soln} shows the accuracy of the predicted solution vs the reference solution.
 
\begin{figure}[H]
	\centering
	\includegraphics[width=\textwidth,height=3cm]{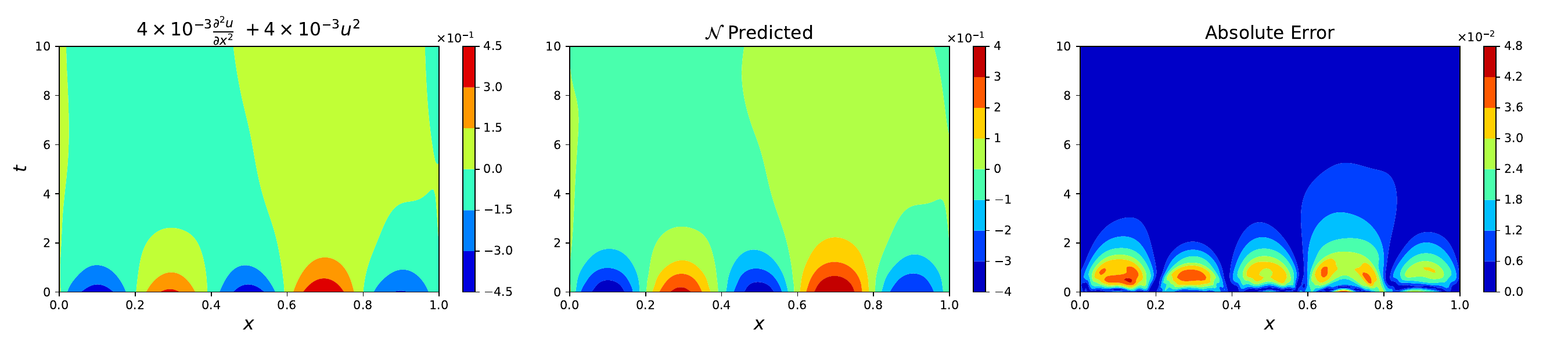}
	\caption{ Comparison of actual hidden PDE terms and output $\mathcal{N}$  for test parameters  $D = 4 \times 10^{-3}$,  $K = 4 \times 10^{-3}$ and input $f$ as given in form  \ref{periodic_input_function}}	
	\label{hidden_eqn_same_kind_parameter_gen_soln}
\end{figure}

The plot in figure \ref{hidden_eqn_same_kind_parameter_gen_soln} shows the how the model has learned the hidden terms of the PDE in comparison to the reference data, in the situation of a combined input and parameter change.
 
.   
 
\subsection{Prediction for unseen parameter and out of distribution input functions}
In this subsection we check the capacity of the model to handle a combination of parameter and input change. Here, the inputs considered are of different type other than the random periodic ones used for training.
 \subsubsection{Unseen parameter and unseen quadratic input}
First, we predict the solution for quadratic input function $f=x^2-x$  along with $D = 4 \times 10^{-3}$ and $K = 4 \times 10^{-3}$. 

\begin{figure}[H]
	\centering
	\includegraphics[width=\textwidth,height=3cm]{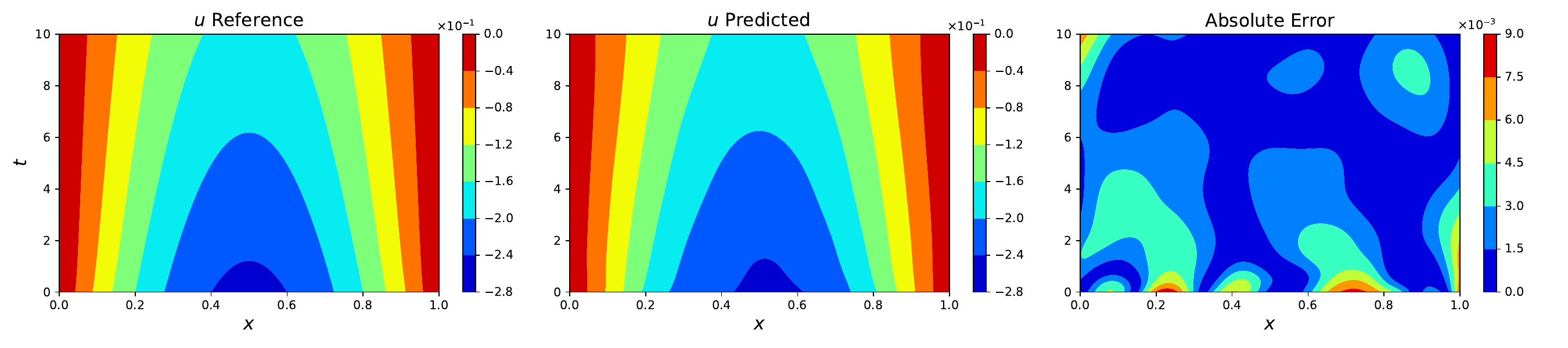}
	\caption{Reference vs predicted $u$ for parameter $D = 4 \times 10^{-3}$, $K = 4 \times 10^{-3}$ and  input function $f=x^2-x$,  $L_2 \  \text{Error} \ 1.45e-2$ }
	\label{prediction_quadratic_parameter_gen_soln}
\end{figure}.

\begin{figure}[H]
	\centering
	\includegraphics[width=\textwidth,height=3cm]{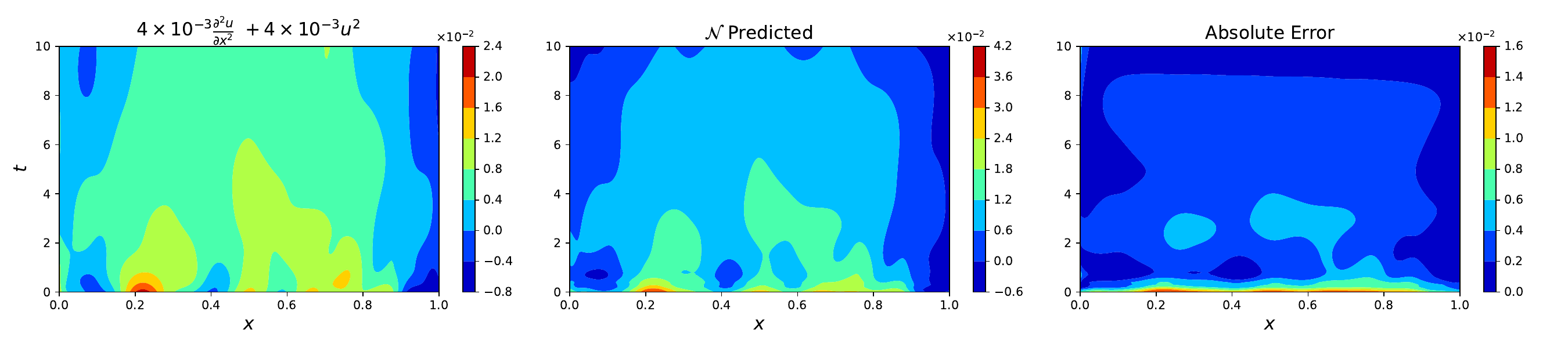}
	\caption{Comparison of actual hidden PDE terms and output $\mathcal{N}$ for parameters $D = 4 \times 10^{-3}$,  $K = 4 \times 10^{-3}$ and  input function $f=x^2-x$ }	
	\label{hidden_eqn_quadratic_parameter_gen_soln}
\end{figure}.
 
 \subsubsection{Unseen parameter and unseen cubic input} 

Next, we consider a cubic test input function $f=x^3-1.5x^2+0.5x$ with $D = 4 \times 10^{-3}$ and $K = 4 \times 10^{-3}$. 

\begin{figure}[H]
	\centering
	\includegraphics[width=\textwidth,height=3cm]{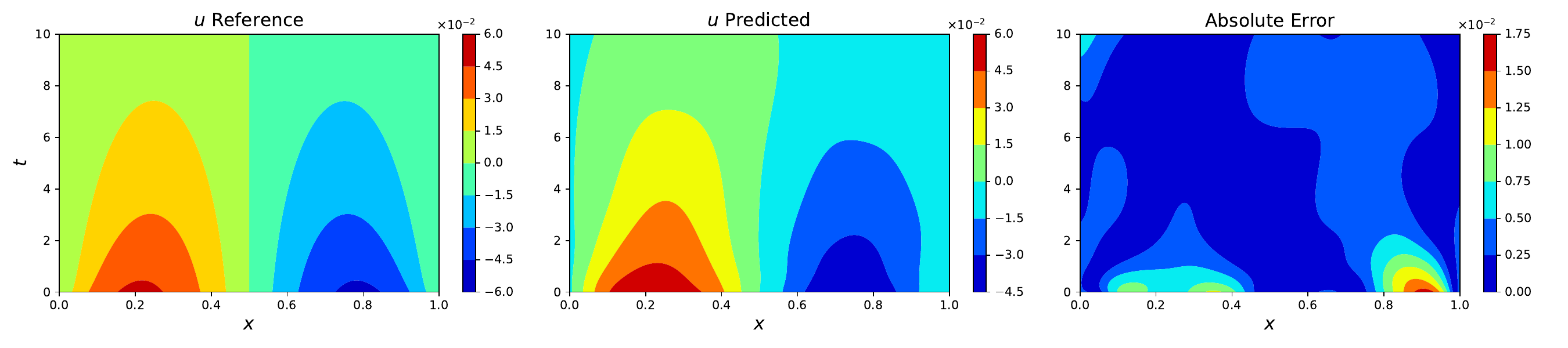}
	\caption{Reference vs predicted $u$ for parameter $D = 4 \times 10^{-3}$,  $K = 4 \times 10^{-3}$ and input function $f=x^3-1.5x^2+0.5x$,  $L_2 \  \text{Error} \ 1.68e-1$ }
	\label{prediction_cubic_parameter_gen_soln}
\end{figure}.

\begin{figure}[H]
	\centering
	\includegraphics[width=\textwidth,height=3cm]{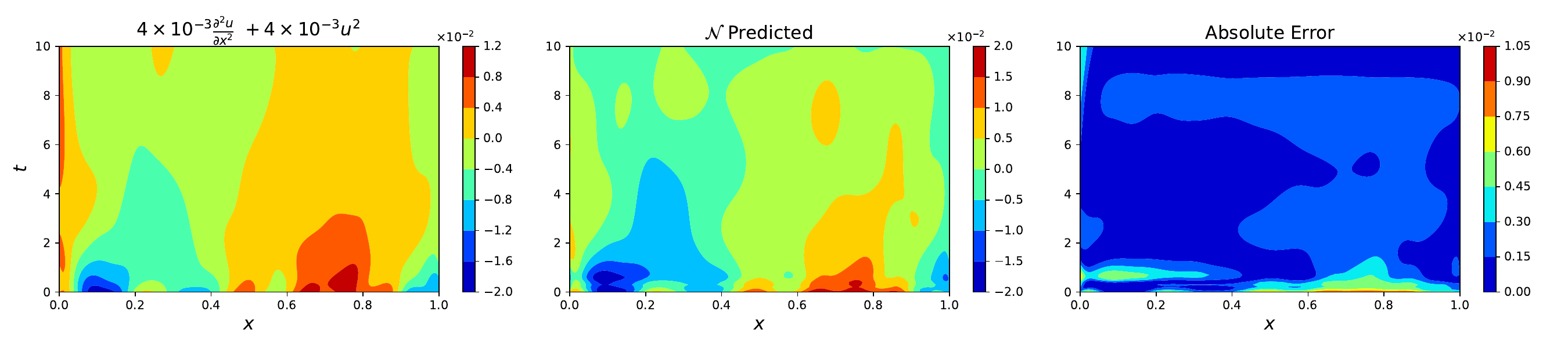}
	\caption{Comparison of actual hidden PDE terms and  output $\mathcal{N}$ for parameter $D = 4 \times 10^{-3}$,  $K =4 \times 10^{-3}$ and  input function $f=x^3-1.5x^2+0.5x$ }	
	\label{hidden_eqn_cubic_parameter_gen_soln}
\end{figure}.

\subsubsection{Unseen parameter and unseen trigonometric input}
Finally, we consider a trigonometric input function given by $f=x-tan(\frac{\pi x}{4})$ with $D = 4 \times 10^{-3}$ and $K = 4 \times 10^{-3}$. 
 
\begin{figure}[H]
	\centering
	\includegraphics[width=\textwidth,height=3cm]{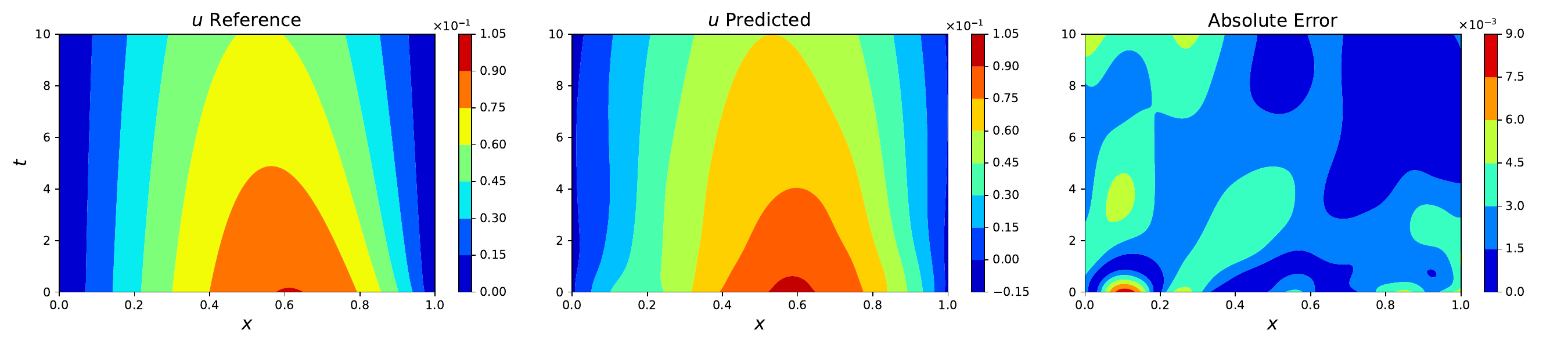}
	\caption{Reference vs predicted $u$ for parameters  $D = 4 \times 10^{-3}$,  $K = 4 \times 10^{-3}$ and input function $f=x-tan(\frac{\pi x}{4})$ , $L_2 \  \text{Error} \ 4.70e-2$ }
	\label{prediction_trignometric_parameter_gen_soln}
\end{figure}. 

 \begin{figure}[H]
 	\centering
 	\includegraphics[width=\textwidth,height=3cm]{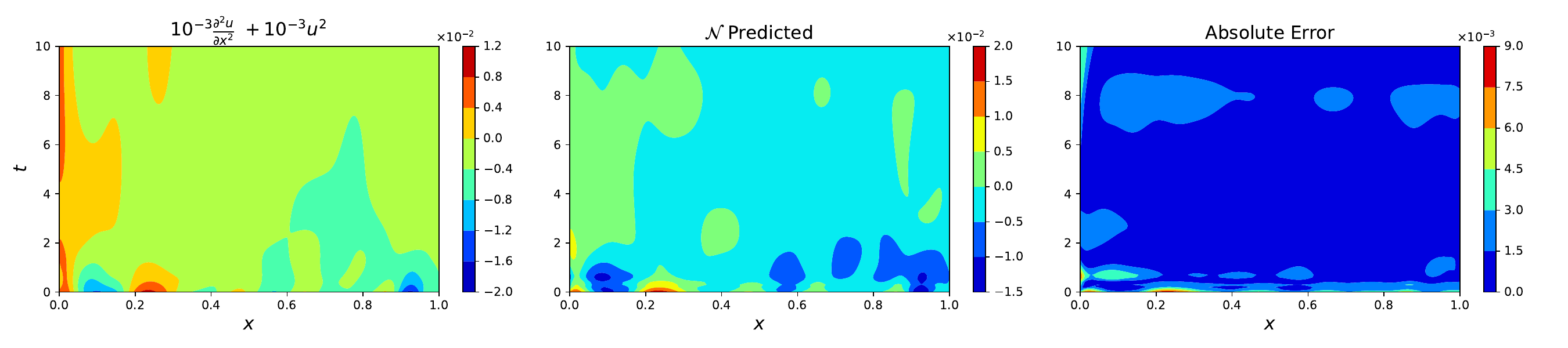}
 	\caption{Comparison of actual hidden PDE terms and  output $\mathcal{N}$ for parameter  $D = 4 \times 10^{-3}$ ,  $K = 4 \times 10^{-3}$ and  input function  $f=x-tan(\frac{\pi x}{4})$ }	
 	\label{hidden_eqn_trigonometric_parameter_gen_soln}
 \end{figure}.

\subsection{Error analysis for different kind of input functions}
\begin{figure}[H]
	\centering
	\begin{subfigure}[b]{0.475\textwidth}
		\centering
		\includegraphics[width=\textwidth]{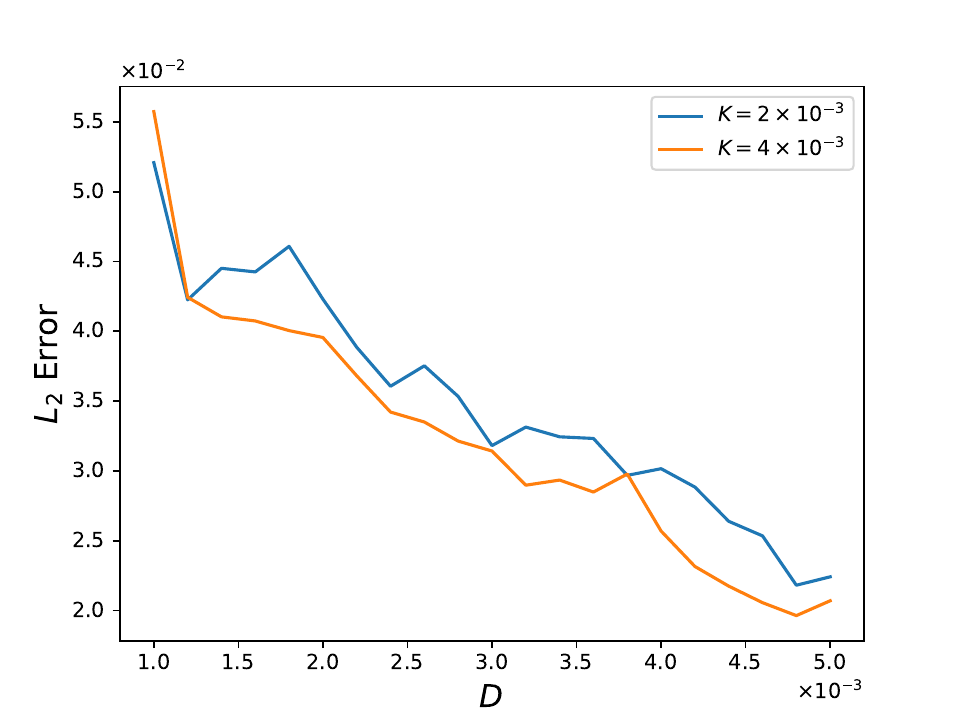}
		\caption{Mean $L_2$ Error for prediction with 100 random input functions over parameters}  
		\label{l2_error_param_same_kind}
	\end{subfigure}
	\hfill
	\begin{subfigure}[b]{0.475\textwidth}  
		\centering
		\includegraphics[width=\textwidth]{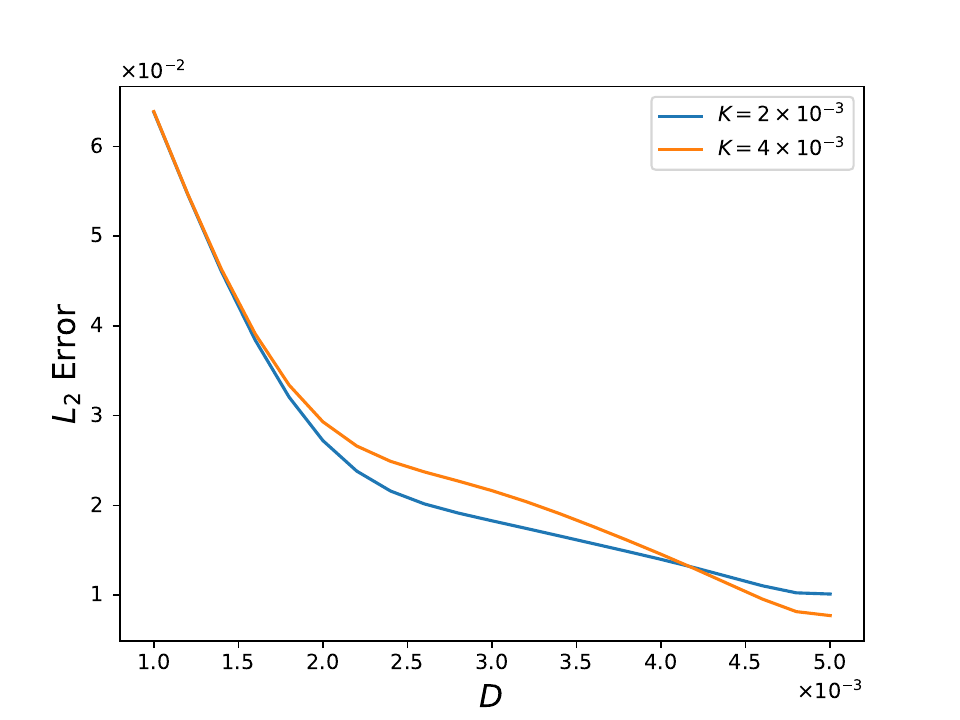}
		\caption{$L_2$ Error for prediction with quadratic input function over the parameters }	
		\label{l2_error_param_quadratic}
	\end{subfigure}
	\vskip\baselineskip
	\begin{subfigure}[b]{0.475\textwidth}   
		\centering
		\includegraphics[width=\textwidth]{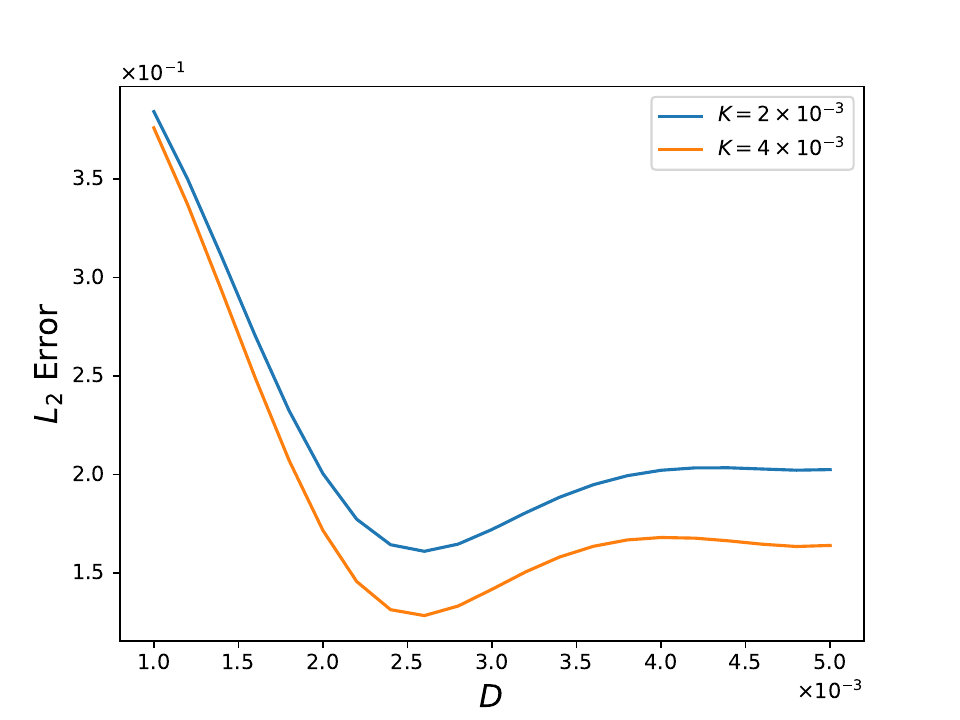}
		\caption{Cubic input function }	
		\label{l2_error_param_cubic}
	\end{subfigure}
	\hfill
	\begin{subfigure}[b]{0.475\textwidth}   
		\centering
		\includegraphics[width=\textwidth]{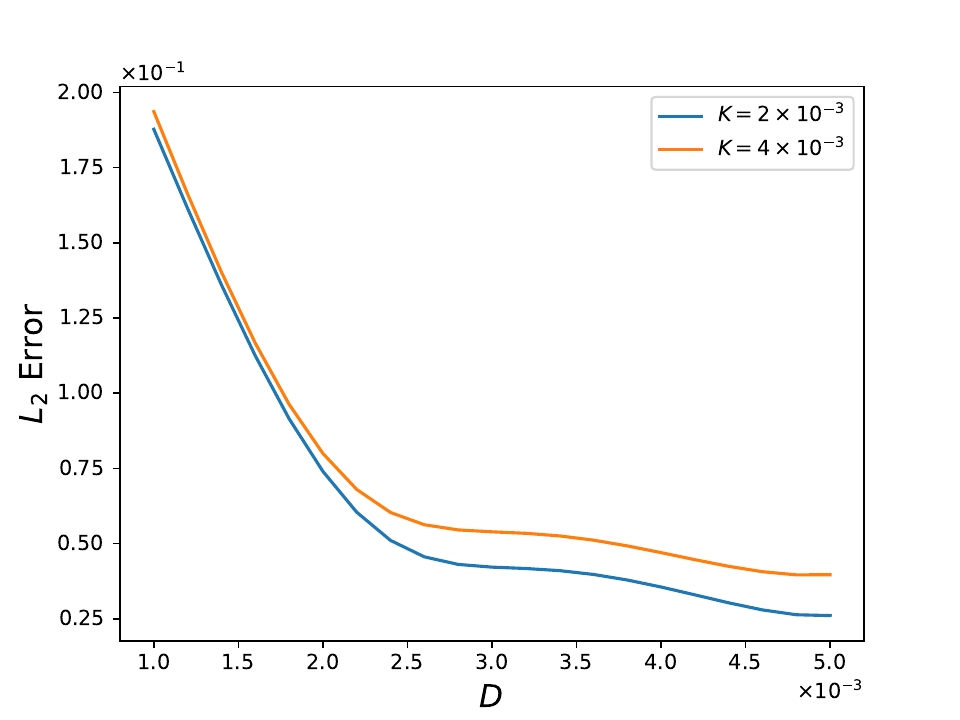}
		\caption{Trigonometric input function }	
		\label{l2_error_param_trignometric}
	\end{subfigure}
	\caption{$L_2$  Error over the test parameter for different kind of input functions}
	\label{l2_error_param}
\end{figure}

To analyze the error and prediction quality over a range of parameter values we predict the solution for test parameter $D$  at 21 equispaced points in range $10^{-3}$ to $5 \times 10^{-3}$  and for each value of $K$ in the set $ \{2 \times 10^{-4}, 4 \times 10^{-4} \}$. The plot in Figure \ref{l2_error_param_same_kind} shows the mean $L_2$ Error of predictions with 100 random input functions. The Figure in   \ref{l2_error_param_quadratic} ,   \ref{l2_error_param_cubic},   \ref{l2_error_param_trignometric} shows $L_2$ Error over different parameters for quadratic, cubic and trigonometric input function respectively.

As we can observe in Figure \ref{l2_error_param}, for all different kind of input functions, the error consistently decreases as the system becomes more diffusion dominated.

Here, in the above examples we can observe that even with only the time derivative known, the model is able to learn the effect of combined parameter and input change. It's even more interesting because parameters $D$ and $K$ are the coefficients of PDE terms, hence the effect of these terms were learnt by the network $N_{hid}$ in the form of a black box, and was able to make good predictions of not only the state but also able to learn decently the hidden dynamics of the system. This we beleive is due to the incorporation of the available physical knowledge into the system.

\section{Domian generalization}
 Domain change or domain deformation is another common phenomenon in many scientific and industrial applications. Depending on application, the deformation in domain can range from a few microns to large deformations leading to complete change in shape and size of the system. It's highly pertinent for design and development applications, where at each design stage, the shape and size of the domain gets slightly modified. However, for most applications, an expert might have an estimate on how much elongation or contraction the domain might undergo over it's extended period of use. The aim here is to learn the system dynamics corresponding to different domain configuration within a certain range. Once learned, this would mean that for every small change in domain one doesn't need to retrain the model. This can end up saving a lot of effort and resources.  Here, in this section, we shall present an extension of the network architecture shown in figure \ref{HPM_gen_input_net_archt} to handle a combination of input and domain change in the system. 

\begin{figure}[H]
	\centering
	\includegraphics[width=6cm]{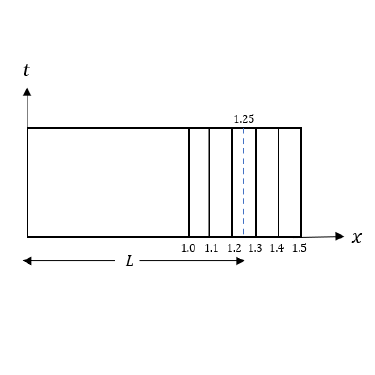}
	\caption{The deformation of the domain along x with different domain configurations }	
	\label{domain_deform}
\end{figure}

To keep things consistent with our study, here too we will consider the same RD system described in equation  \ref{reaction_diffusion_eqn}. The domain under consideration is $x \in [0,L]$ and $t \in [0,10]$. Here $L$ is the domain length. We assume that the domain undergoes elongation only along the x-axis i.e, the length $L$ varies from 1 to 1.5 unit, with $L=1$ unit corresponding to the initial undeformed state to the maximum of $L=1.5$ units in the fully deformed state. Our objective here is to train the model for different domain configuration within the range of $ L \in [1,1.5]$. For training we consider 6 different lengths $L$  in the set $\{1,1.1,1.2,1.3,1.4,1.5 \}$. Further, after training we aim to predict at an intermediate value of $L$. Schematically these domain configurations has been shown in figure \ref{domain_deform}.

\begin{figure}[H]
	\centering
	\includegraphics[width=4cm]{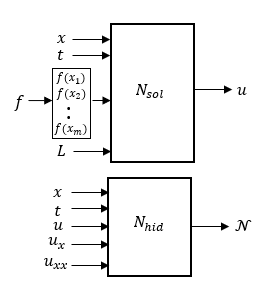}
	\caption{Network architecture for domain length and input function generalization}	
	\label{hpm_net_arch_domain}
\end{figure}

\begin{figure}[H]
	\centering
	\includegraphics[width=10cm]{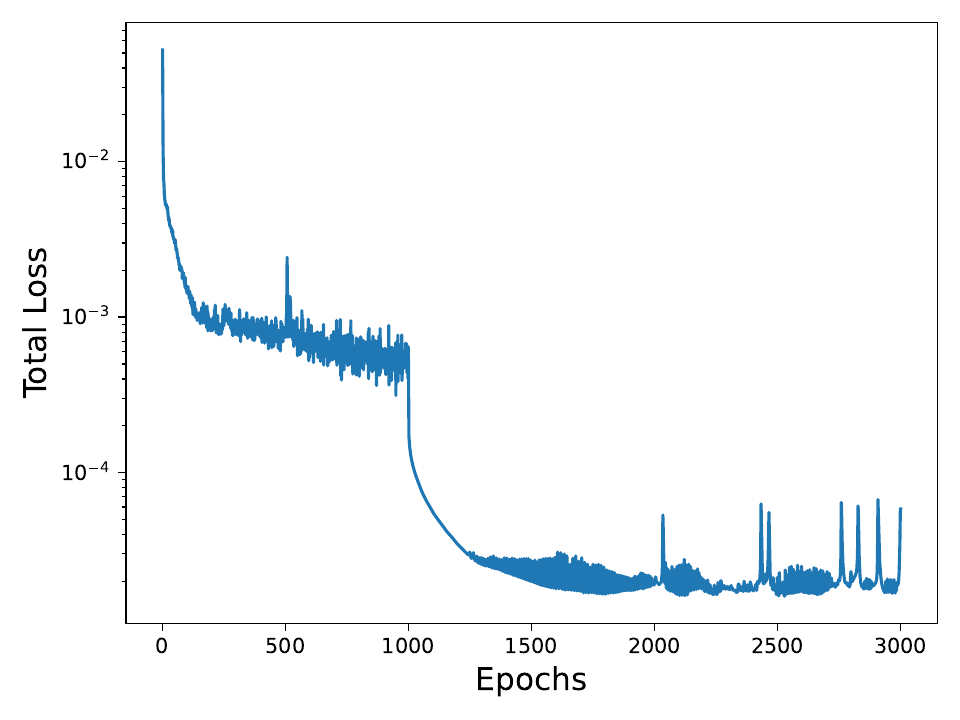}
	\caption{ Total Loss over the training }	
	\label{total_loss_domain_plot}
\end{figure}

Here too, we consider the RD system as represented in equation \ref{reaction_diffusion_eqn}, with just the time derivative term known. We enrich the feature space of the solution network  as shown in figure \ref{HPM_gen_input_net_archt} by inclusion of the domain length $L$ as an additional feature. The schematic for the networks are shown in Figure \ref{hpm_net_arch_domain}. The model is trained with 200 random input functions of the form \ref{periodic_input_function} for each considered domain configuration, and $n_{data} = 500$. The considered training data is approximately $2.5\%$  of entire reference data. The model has been trained using Adam's optimization with a learning rate of $10^{-3}$ for the first 1000 epochs and then for another 2000 epochs with a modified learning rate of $10^{-4}$. Moreover, we consider 1000 collocation points at each gradient descent step. In figure \ref{total_loss_domain_plot} we see that the training loss decays over the epochs, suggesting convergence. 

\subsection{Unseen domain configuration and similar kind of unseen input function}  

\begin{figure}[H]
	\centering
	\includegraphics[width=\textwidth,height=3.5cm]{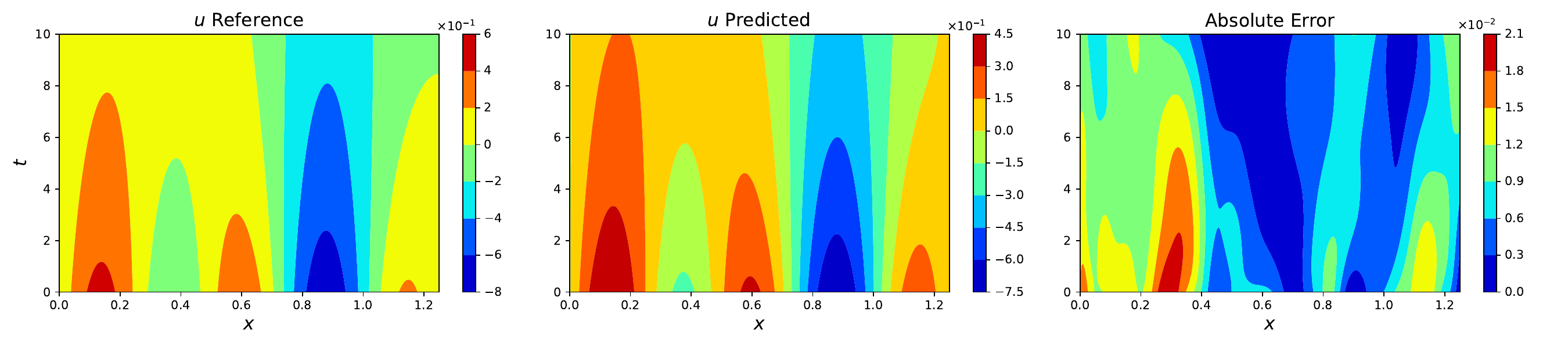}
	\caption{Reference vs predicted $u$ for L=1.25 and random periodic input function given by the form in \ref{periodic_input_function},   $L_2 \  \text{Error} \ 3.58e-2$ }
	\label{random_periodic_input_dom_gen}
\end{figure}

We first check the proposed method in the case of a similar random periodic input function for an unseen domain configration $L=1.25$ . The contour in \ref{random_periodic_input_dom_gen} shows the comparison of actual and model predicted solution. 
 .

Morover, as done throughout the study, we check the model's ability in learning the hidden physics of the system. The plot in figure \ref{hidden_eqn_same_kind_parameter_gen_soln} compares the actual hidden PDE terms with learned $\mathcal{N}$.

 \begin{figure}[H]
	\centering
	\includegraphics[width=\textwidth,height=3.5cm]{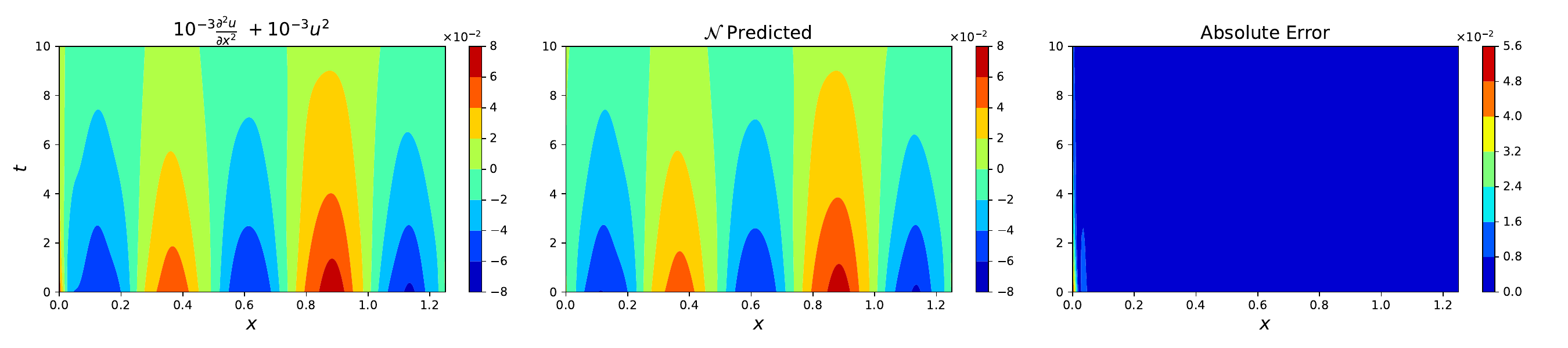}
	\caption{Comparison of actual hidden PDE terms and learned $\mathcal{N}$ for $ L=1.25$ and random  periodic input function $f$ given by the form in \ref{periodic_input_function}}	 
	\label{random_periodic_input_hidden_dom_gen}
\end{figure}.

\subsection{Unseen domain configuration and out of distribution input functions}  
Here, we check the capacity of the model to handle a combination of unseen domain configuration as well as unseen type of input functions.

\subsubsection{Quadratic input function}  
First, we consider a quadratic input function $f=x(x-\text{L})$ with $L=1.25$. The corresponding plots are shown below. 
\begin{figure}[H]
	\centering
	\includegraphics[width=\textwidth,height=3.5cm]{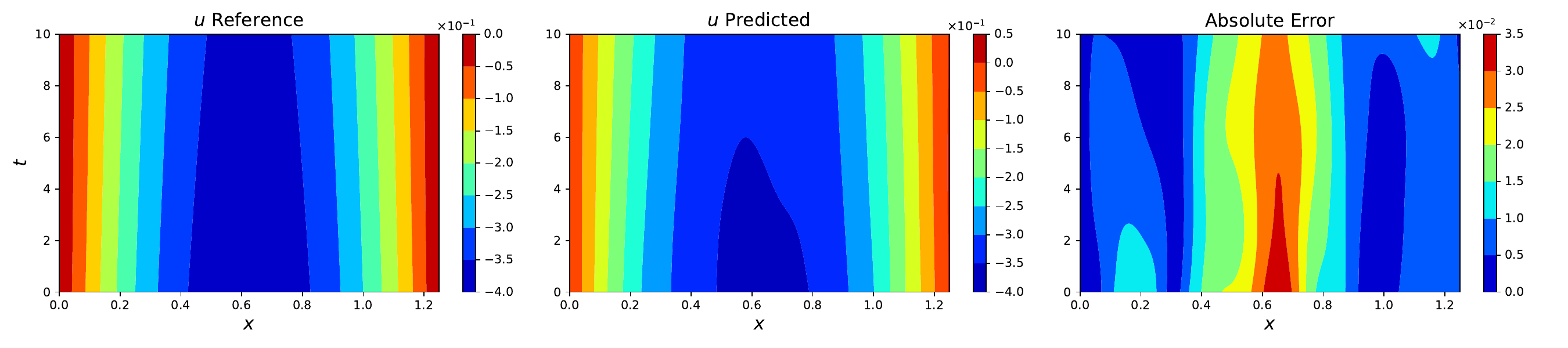}
	\caption{Reference vs predicted $u$ for $ L=1.25$ and quadratic input function $f=x(x-\text{L})$, $L_2 \  \text{Error} \ 5.23e-2$ }
	\label{quadratic_input_dom_gen}
\end{figure}.

\begin{figure}[H]
	\centering
	\includegraphics[width=\textwidth,height=3.5cm]{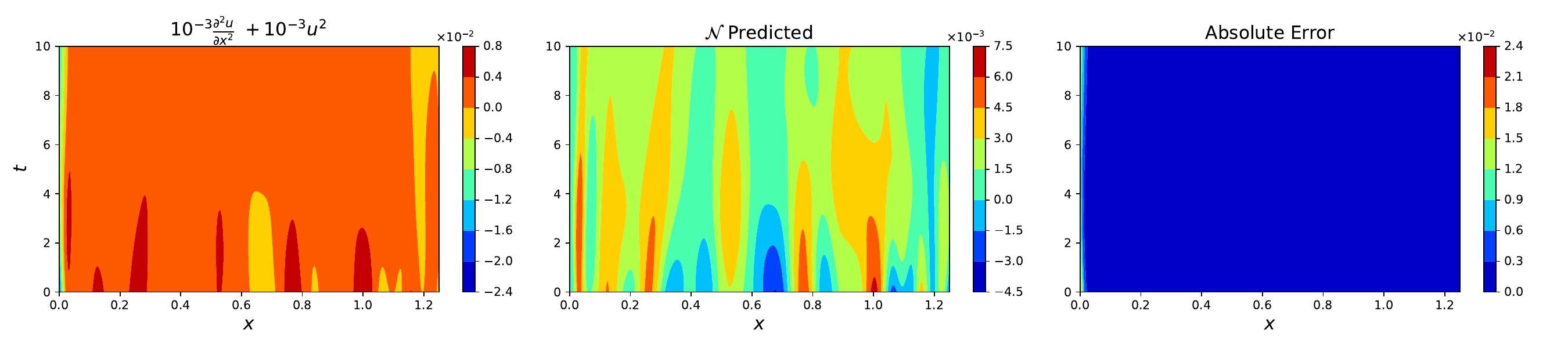}
	\caption{Comparison of actual hidden PDE terms and learned $\mathcal{N}$ for $ L=1.25$ and  quadratic input function $f=x(x-\text{L})$}	
	\label{quadratic_input_hidden_dom_gen}
\end{figure}.

\subsubsection{Cubic input function}  
Next, we predict for cubic input function $f=x(x-\text{L})(x-\text{L}/2)$ with $L=1.25$. The corresponding plots are shown below.

\begin{figure}[H]
	\centering
	\includegraphics[width=\textwidth,height=3.5cm]{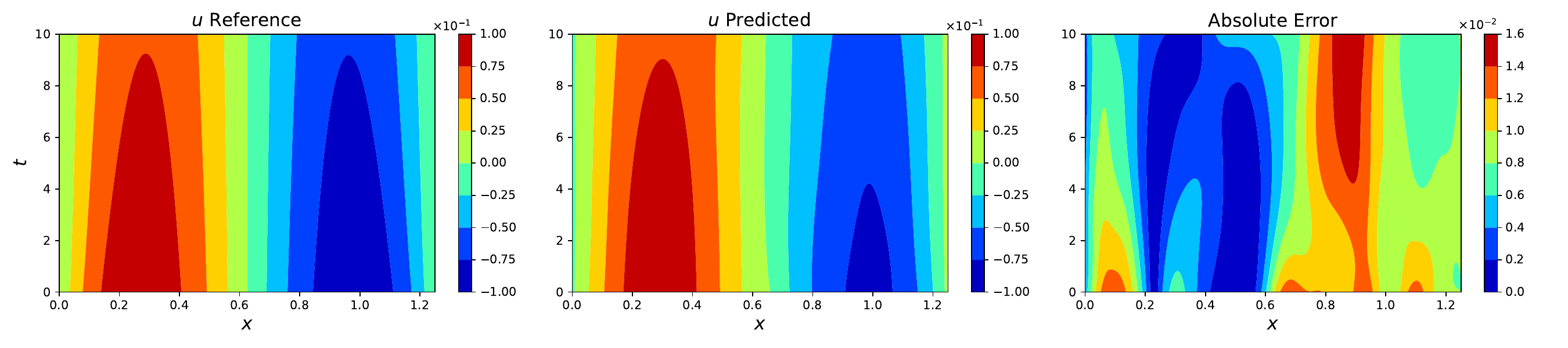}
	\caption{Reference vs predicted $u$ for $L=1.25$ and cubic input function $f=x(x-\text{L})(x-\text{L}/2)$,  $L_2 \  \text{Error} \ 1.35e-1$ }
	\label{cubic_input_dom_gen}
\end{figure}.
 
\begin{figure}[H]
	\centering
	\includegraphics[width=\textwidth,height=3.5cm]{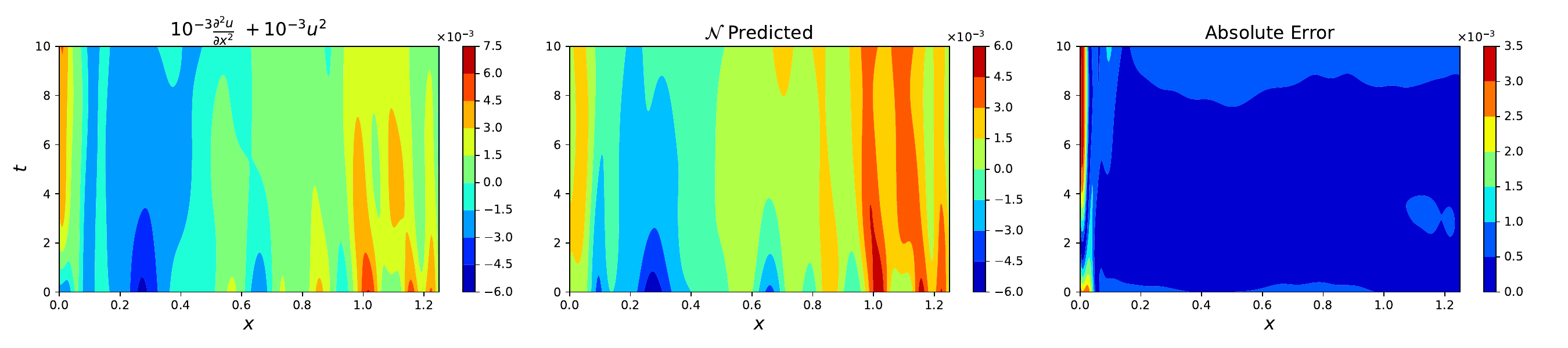}
	\caption{Comparison of actual hidden PDE terms and the learned $\mathcal{N}$ for $ L=1.25$ and cubic input function $f=x(x-\text{L})(x-\text{L}/2)$}	
	\label{cubic_input_hidden_dom_gen}
\end{figure}.

\subsubsection{ Trigonometric input function}  
Finally, we predict for trigonometric input function $f=\frac{x}{L}-tan(\frac{\pi x}{4L})$ with $L=1.25$. The corresponding plots are shown below.
\begin{figure}[H]
	\centering
	\includegraphics[width=\textwidth,height=3.5cm]{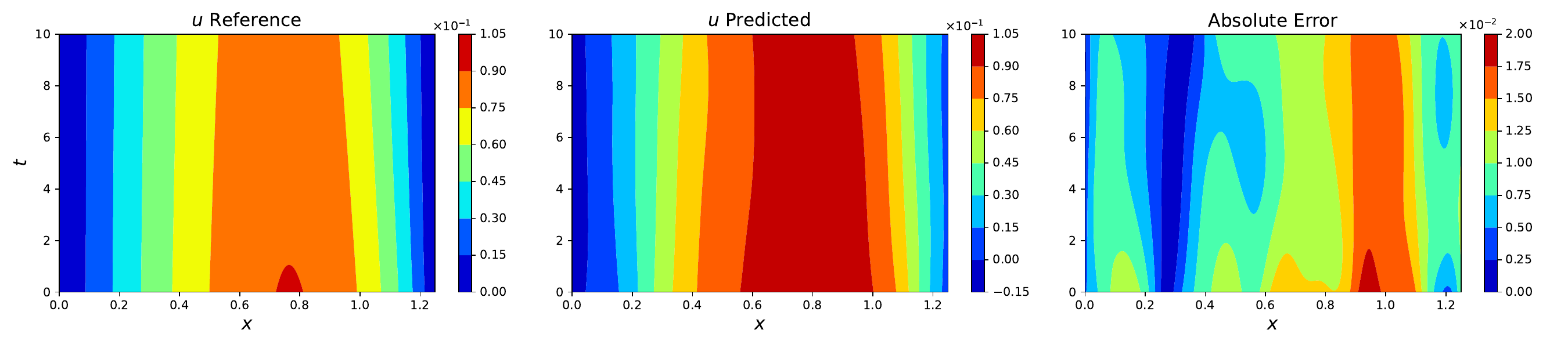}
	\caption{Reference vs predicted $u$ for $ L=1.25$ and trigonometric input function $f=\frac{x}{L}-tan(\frac{\pi x}{4L})$,   $L_2 \  \text{Error} \ 1.65e-1$ }
	\label{trigono_input_dom_gen}
\end{figure}.

\begin{figure}[H]
	\centering
	\includegraphics[width=\textwidth,height=3.5cm]{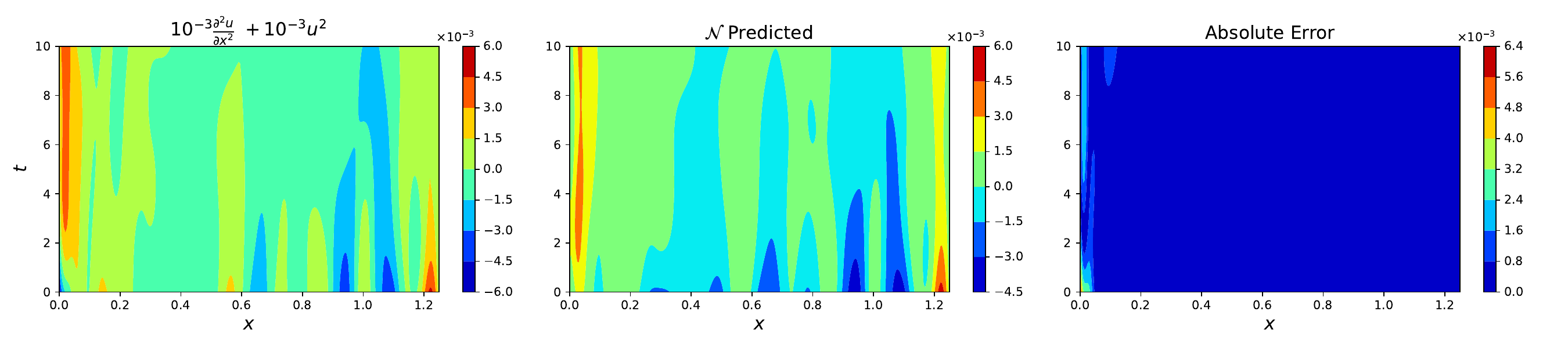}
	\caption{Comparison of actual PDE terms and learned $\mathcal{N}$ for $ L=1.25$ and trigonometric input function $f=\frac{x}{L}-tan(\frac{\pi x}{4L})$}	
	\label{trigono_input_hidden_dom_gen}
\end{figure}.

The results above shows that the proposed methodology holds promise in model generalization for a combination of inputs and domain change. 

\section{Conclusion and discussion}

In this paper we have proposed a methodology and introduced a novel framework based on hidden physics models which is applicable for modelling systems where the full governing equations are unknown. The framework consists of a pair of neural networks, where one network is used to represent the unknown state and the second to represent the hidden physics. The proposed methodology  has three main aspects. First, the framework has been shown to handle situations where the governing equations are not fully known. Second, it has demonstrated ability to react to system change like inputs, parameters, domain configuration and combinations of them. Third, the second network in the proposed framework shows promise in learning the hidden physics in an implicit/black box way. A system of reaction diffusion equations has been considered in this study, with only the time derivative term as known, and extensively studied for different unseen input types, parameters and domain configuration. Thorough description of the training and analysis of the test results have been provided, to showcase the efficacy of the method. Though neural networks are very flexible and robust in learning different kind of systems, it's solution suffers from lack of interpreatability. Moreover, given enough data a network might just memorize the data and can only make decent predictions for test data which is of similar kind as those used in training. This issue of overfitting is very relevant for model deployment. In our proposed methodolgy, the different kind of unseen input functions considered plus the learned hidden terms of the PDE holds promise in demonstrating that the networks doesn't just memorize but learns the underlying system dynamics. 
Another observation is that the proposed methodology can help in discovering the physics of system based on multiple datasets (in an implicit form). Here, we have datasets corresponding to different initial conditions for the same differential equation. Hence, we are able to learn the hidden terms of the PDE from different datasets. Once the hidden dynamics is learned accurately in the form of black box, one can use the learned equations to solve for different geometry and also extrapolate in time using PINNs. However, if the model is trained with just a single dataset then the model may have bias towards that particular training dataset and may not learn unique equations \cite{RaissiDHP}. Therefore training with multiple dataset simultaneously results in more accurate learning of the hidden terms of the PDE.

The methodology and the model framework proposed here is based on a pair of neural networks. We chose a neural network based approach due to it's simplicity and easy implementation. The very same idea shown in this paper can be implemented with gaussian processes as well.

\section*{References}

\bibliography{DHPM_generalization}

\end{document}